%% file: main.tex
\documentclass[conference]{IEEEtran}
\IEEEoverridecommandlockouts
\usepackage{cite}
\usepackage{amsmath,amssymb,amsfonts}
\usepackage{algorithmic}
\usepackage{graphicx}
\usepackage{textcomp}
\usepackage{xcolor}
\usepackage[table]{xcolor}
\usepackage{multirow}
\usepackage{multicol}
\usepackage{booktabs}
\usepackage{enumitem}
\usepackage{amsfonts}
\usepackage{symbols}
\usepackage{hyperref}
\usepackage{geometry}
\usepackage{float}
\usepackage{placeins}

\newcommand{\best}[1]{\cellcolor{red!25}\textbf{#1}}
\newcommand{\second}[1]{\cellcolor{yellow!25}#1}

\def\BibTeX{{\rm B\kern-.05em{\sc i\kern-.025em b}\kern-.08em
    T\kern-.1667em\lower.7ex\hbox{E}\kern-.125emX}}

\begin{document}

\newgeometry{top=2.54cm, left=1.91cm, right=1.91cm, bottom=1.91cm}

\title{OctoSplat: Hybrid OctoMap–Gaussian Splatting for Active Semantic Mapping and Phenotyping with Horticultural Robots\\
}

\author{Jose Cuaran$^{1}$, Naveen Kumar Uppalapati$^{3}$, and Girish Chowdhary$^{1,2}$% <-this % stops a space

\thanks{The authors are with (1) the Siebel School of Computing and Data Science, (2) the Department of Agricultural and Biological Engineering  and (3) National Center for Supercomputing Applications at University of Illinois, Urbana-Champaign.
        }%
\thanks{{Correspondence to \tt\small \{jrc9,girishc\}@illinois.edu}}
}

\maketitle

%\begin{figure*}[!t]  % top of the page, can span both columns
%    \centering
%    \includegraphics[width=\textwidth]{figures/Overview Figure.png}
%    \caption{System Overview. (a) The user prompts high level monitoring task from the robot. The robot breaks down the task into required action sequence such as locomotion and arm manipulation along with corresponding tool calls. (b) The robot uses sensor data to check current state of the manipulator and robot pose in the map. (c) GPT-4.1 uses Semantic map to pinpoint a goal for the locomotion task tool. (d) Tool call used to identify and center object in camera frame for the manipulator arm for monitoring. }
%    \label{fig:overview}
%\end{figure*}

\begin{abstract}
Semantic reconstruction of agricultural scenes plays a vital role in tasks such as phenotyping and yield estimation. However, traditional approaches that rely on manual scanning or fixed camera setups remain a major bottleneck in this process, and existing active-mapping methods based solely on occupancy grids are too coarse to support accurate trait estimation. To close this gap, we propose an active 3D reconstruction framework for horticultural environments using a mobile manipulator. The proposed system integrates the classical OctoMap representation with 3D Gaussian Splatting to enable accurate and efficient target-aware mapping. While a low-resolution OctoMap provides probabilistic occupancy information for informative viewpoint selection and collision-free planning, 3D Gaussian Splatting leverages geometric, photometric, and semantic information to optimize a set of 3D Gaussians for high-fidelity scene reconstruction. We further introduce a robust mapping strategy that mitigates the effects of semantic segmentation and depth noise, together with a background pruning method that reduces memory consumption and computational cost. We validate our framework across simulated, laboratory, and real greenhouse scenes, and show that the improvements hold consistently across three state-of-the-art Gaussian Splatting backbones. In simulation, where ground-truth geometry is available, our approach outperforms occupancy-based mapping in both reconstruction accuracy and runtime efficiency: compared with a 0.01\,m-resolution OctoMap, our method doubles the fruit-level F1 score under noisy conditions while achieving up to a threefold reduction in runtime. Beyond simulation, novel-view synthesis quality also improves consistently in the laboratory and real greenhouse environments, with PSNR and mIoU improving by up to 1.5 dB and 18\% respectively. Finally, the reconstructed semantic maps enable fruit counting and volume estimation with accuracies approaching 80\%.\\
\textbf{Supplementary material}: \href{https://jrcuaranv.github.io/octosplat/}{https://jrcuaranv.github.io/octosplat/}.
\end{abstract}


\section{Introduction}
\input{sections/introduction}

\FloatBarrier  % ensures all previous figures appear before this section

\vspace{-0.1cm}
\section{Related works}
\label{sec:relatedworks}
\input{sections/related_works}

\vspace{-0.1cm}
\section{Methods}
\label{sec:methods}
\input{sections/methods}

\vspace{-0.2cm}
\section{Experiments}
\label{sec:experiments}
\input{sections/experiments}

\section{Results}
\label{sec:results}
\input{sections/results}

\FloatBarrier  % ensures all previous figures appear before this section

\section{Discussion}
\label{sec:discussion}
\input{sections/discussion}

\vspace{-0.2cm}
\section{Conclusion}
\input{sections/conclusion}

\bibliographystyle{IEEEtran}
\bibliography{mybibliography}

\section{Appendix}
\input{sections/appendix}

\end{document}

%% file: sections/introduction.tex
\begin{figure*}[ht]
\centerline{\includegraphics[width=150mm]{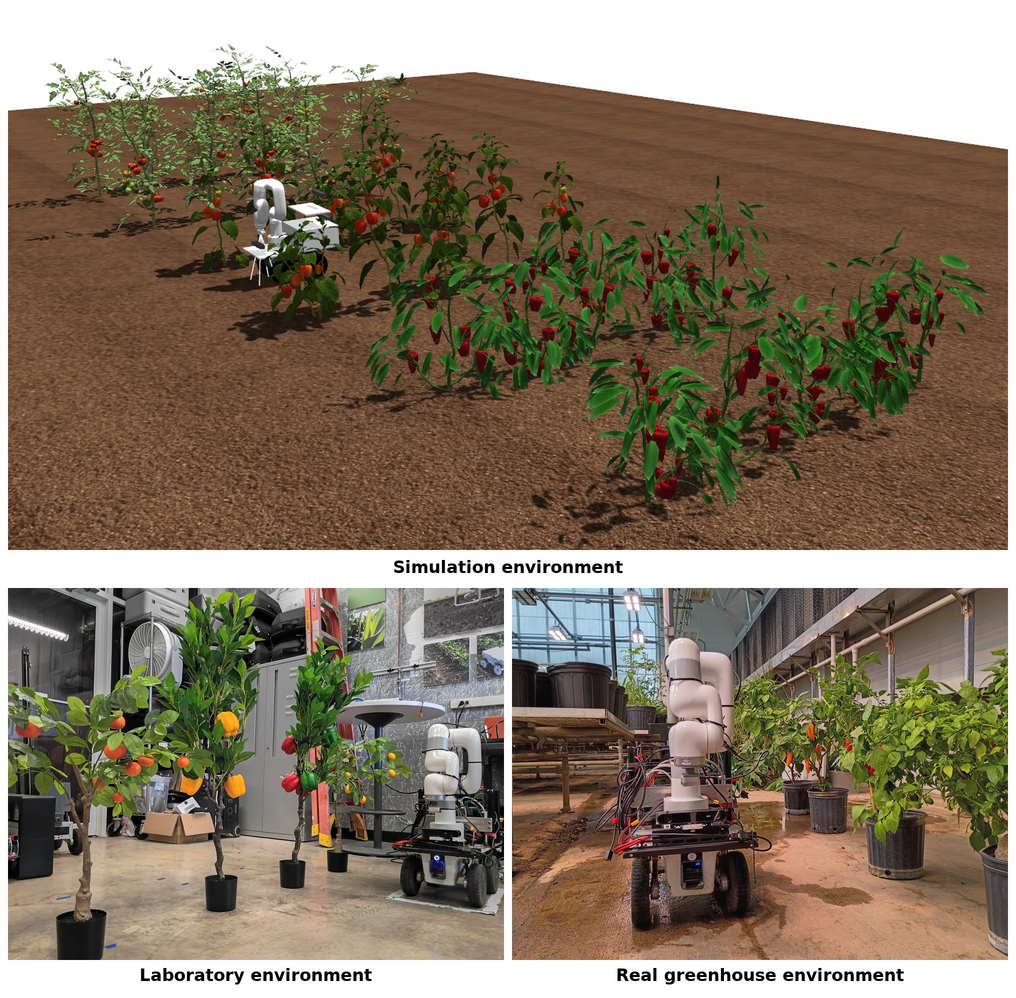}}
\caption{Simulation, laboratory, and real greenhouse environments, along with our simulated and physical mobile manipulator.}
\label{fig:merged_environments}
\vspace{-0.1cm}
\end{figure*} 

Modern agriculture increasingly relies on automation and data-driven technologies to address challenges such as labor shortages, sustainability demands, and the need for improved productivity \cite{padhiary2025emerging}. Within this context, agricultural phenotyping plays a fundamental role in plant breeding, driving advances in yield, resilience, and resource efficiency. At the core of phenotyping tasks lies the collection of image data, which is subsequently used for the 3D reconstruction of plants. These 3D plant models enable the accurate estimation of phenotypic traits such as plant height, fruit size, and biomass content. However, generating high-quality 3D plant models remains a labor-intensive and time-consuming process due to challenges such as occlusions, varying lighting conditions, and the need for precise camera positioning \cite{harandi2023make, maraveas2024image}. Traditional methods often rely on scanning plants with handheld cameras or using fixed-camera setups in controlled environments \cite{dong2020semantic, wang2025p3dfusion}. Robotic platforms have emerged as an alternative for automating data collection in agricultural fields, but most still depend on fixed or predefined camera configurations \cite{dong20174d, pan2023panoptic}.

To overcome these limitations, several studies have explored active mapping methods for agricultural environments \cite{cuaran2025active, pan2023panoptic, freeman2024autonomous, lehnert20193d, zaenker2021viewpoint}, in which the camera pose is continuously adjusted to capture specific targets, such as fruits or leaves. Despite their effectiveness in viewpoint planning, most of these approaches rely on OctoMap-based representations. While OctoMap provides a probabilistic occupancy model that is well suited for information-driven exploration and collision-free motion planning, it often produces low-quality reconstructions and inaccurate volume estimates due to its dependence on voxel resolution \cite{zaenker2021viewpoint}. This limitation is particularly consequential for phenotyping, where downstream traits such as fruit volume and count are computed directly from the reconstructed geometry: coarse voxelization can bias these estimates even when viewpoint planning itself succeeds. Increasing the voxel resolution can improve reconstruction accuracy but comes at a substantially higher computational cost because of the large number of ray-casting operations required. Furthermore, OctoMap cannot be efficiently implemented on GPUs because of its hierarchical tree structure, limiting its scalability and real-time performance \cite{min2023octomap}.

Recently, neural scene representations such as Neural Radiance Fields (NeRF) \cite{mildenhall2021nerf} and 3D Gaussian Splatting (3DGS) \cite{kerbl3Dgaussians} have emerged as powerful techniques for high-quality novel-view synthesis and 3D reconstruction by exploiting geometric and photometric information. Unlike NeRF, 3DGS provides an explicit scene representation that enables real-time rendering and efficient optimization, making it attractive for capturing the fine-grained geometric detail (e.g., individual fruit surfaces) required for accurate phenotyping. However, 3DGS does not explicitly represent unknown or unobserved space, which is essential for information-driven active mapping and collision-free motion planning.

To address this limitation, we propose a hybrid active semantic mapping framework that combines the complementary strengths of OctoMap and 3DGS. A low-resolution OctoMap provides probabilistic occupancy information for informative viewpoint selection and collision-free motion planning, while 3DGS jointly exploits geometric, photometric, and semantic cues to produce high-fidelity scene reconstructions. This design enables efficient exploration while preserving the reconstruction quality needed for downstream phenotypic trait estimation, such as fruit volume and count, even under realistic sensing noise. To improve robustness, we introduce a mapping strategy that mitigates the effects of noisy depth measurements and imperfect semantic segmentation, together with a background pruning strategy that substantially reduces the memory footprint and computational cost of 3DGS. We validate the proposed framework across 30 simulated, 10 laboratory, and 10 real greenhouse scenes, demonstrating improved reconstruction accuracy and runtime efficiency, as well as the applicability of the resulting reconstructions to fruit volume and count estimation. Furthermore, we evaluate three state-of-the-art Gaussian Splatting backbones and observe consistent improvements across all of them, suggesting that our approach is not tied to a specific architecture. All code and data are publicly available to facilitate reproducibility.

In summary, to achieve this objective, the main contributions of this paper are:
\begin{itemize}
    \item A novel hybrid framework for active semantic mapping that combines a low-resolution OctoMap with 3D Gaussian Splatting for efficient and accurate scene reconstruction.
    \item A robust mapping strategy that mitigates the effects of semantic segmentation errors and depth sensor noise.
    \item A background pruning method that reduces memory consumption and training time while improving reconstruction accuracy.
    \item A demonstration that these strategies generalize across three different Gaussian Splatting backbones and across simulation, laboratory, and real greenhouse environments.
\end{itemize}

% \clearpage
% \restoregeometry

%% file: sections/related_works.tex
\begin{table*}[htbp]
\centering
\caption{Comparison of related works on plant/crop 3D reconstruction and active mapping. Ours combines a low-resolution semantic OctoMap for viewpoint planning and collision-free navigation with semantic 3D Gaussian Splatting for high-fidelity, target-aware reconstruction.}
\label{tab:related_works}
{
\resizebox{\textwidth}{!}{%
\begin{tabular}{@{}llllllll@{}}
\toprule
Method & Scene & Domain & Active & Target & Sem. & BP & Val. \\
 & Representation &  &  & aware &  &  &  \\
\midrule
\cite{dong20174d} & PC (4D) & Agri & $\times$ & $\times$ & $\times$ & $\times$ & Real \\
\cite{dong2020semantic} & PC/SfM & Agri & $\times$ & $\times$ & $\times$ & $\times$ & Real \\
\cite{gao2021canopy} & LiDAR PC & Agri & $\times$ & $\times$ & $\times$ & $\times$ & Real \\
\cite{zhang2025wheat3dgs} & 3DGS & Agri & $\times$ & $\times$ & $\times$ & $\times$ & Real \\
\cite{stuart2025high} & 3DGS/NeRF & Agri & $\times$ & $\times$ & $\times$ & $\times$ & Real \\
\cite{mcafee2025evaluation} & 3DGS & Agri & $\times$ & $\times$ & $\times$ & $\times$ & Real \\
\cite{jiang2025cotton3dgaussians} & 3DGS & Agri & $\times$ & $\times$ & $\times$ & $\times$ & Real \\
\cite{shen2026thin} & 3DGS & Agri & $\times$ & $\times$ & $\times$ & $\times$ & Real \\
\cite{ojo2024splanting} & 3DGS & Agri & $\times$ & $\times$ & $\times$ & $\checkmark$ & Real \\
\cite{wang2025p3dfusion} & 3DGS & Agri & $\times$ & $\times$ & $\times$ & $\checkmark$ & Real \\
\cite{chen2025plant} & 3DGS & Agri & $\times$ & $\times$ & $\times$ & $\checkmark$ & Real \\
\cite{li2026object} & 3DGS & Agri & $\times$ & $\times$ & $\times$ & $\checkmark$ & Real \\
\midrule
\cite{burusa2022attention} & Occ. & Agri & $\checkmark$ & $\checkmark$ & $\checkmark$ & $\times$ & Both \\
\cite{zaenker2021viewpoint} & Occ. & Agri & $\checkmark$ & $\checkmark$ & $\checkmark$ & $\times$ & Both \\
\cite{menon2023nbv} & Hybrid (Occ. + TSDF) & Agri & $\checkmark$ & $\checkmark$ & $\checkmark$ & $\times$ & Both \\
\cite{zaenker2023graph} & Occ. & Agri & $\checkmark$ & $\checkmark$ & $\checkmark$ & $\times$ & Both \\
\cite{cuaran2025active} & Occ. & Agri & $\checkmark$ & $\checkmark$ & $\checkmark$ & $\times$ & Both \\
\midrule
\cite{tao2024rt} & 3DGS & Gen. & $\checkmark$ & $\times$ & $\times$ & NA & Both \\
\cite{li2025activesplat} & Hybrid (3DGS+topo.) & Gen. & $\checkmark$ & $\times$ & $\times$ & NA & Both \\
\cite{jin2025activegs} & Hybrid (2DGS+Occ.) & Gen. & $\checkmark$ & $\times$ & $\times$ & NA & Both \\
\cite{jin2024gs} & 3DGS & Gen. & $\checkmark$ & $\times$ & $\times$ & NA & Sim \\
\cite{xu2025hgs} & Hybrid (3DGS+voxel) & Gen. & $\checkmark$ & $\times$ & $\times$ & NA & Both \\
\cite{polyzos2025activeinitsplat} & 3DGS & Gen. & $\checkmark$ & $\times$ & $\times$ & NA & Both \\
\midrule
\textbf{Ours} & Hybrid (Occ.+3DGS) & Agri & $\checkmark$ & $\checkmark$ & $\checkmark$ & $\checkmark$ & Both \\
\bottomrule
\end{tabular}%
}
}
{\raggedright\footnotesize Abbreviations: \\Representation (PC = point cloud; Occ. = occupancy map; 2D/3DGS = 2D/3D Gaussian Splatting; TSDF = Truncated Signed Distance Field);\\Domain (Agri = agriculture/horticulture; Gen. = generic robotics). \\ Active/Passive: Passive methods use fixed, handheld, or pre-planned routes; active methods autonomously select viewpoints. \\Sem.\ = Semantic (building a a semantic map, not just postprocessing with semantic masks) \\BP = Background Pruning (NA = Not applicable to general-purpose scenes, where the background is part of the reconstructed scene). \\Val.\ = Validation (Sim = simulation only; Real = Laboratory or field experiments).\par}
\vspace{-3mm}
\end{table*}

3D reconstruction of plants has emerged as a valuable tool for plant phenotyping, enabling precise quantification of traits such as plant height, leaf area, stem thickness, canopy volume, and fruit size or count \cite{maraveas2024image}. These traits are critical for evaluating plant health, monitoring stress conditions, and supporting breeding programs. 

Approaches for 3D reconstruction in agriculture can broadly be divided into passive and active methods (see summary in Table \ref{tab:related_works}). Passive methods typically rely on fixed camera setups in controlled environments, handheld imaging sensors, or mobile platforms with limited viewpoint control \cite{dong2020semantic, gao2021canopy, dong20174d}. Multiview stereo (MVS) has been widely adopted in this context, enabling the reconstruction of plant structures from overlapping 2D images captured under varying viewpoints \cite{paulus2019measuring}. More recently, learning-based techniques such as neural radiance fields (NeRFs)\cite{mildenhall2020nerf} and 3D Gaussian Splatting\cite{kerbl3Dgaussians} have shown advantages in producing smooth and photorealistic reconstructions, with improved handling of complex geometries such as thin leaves and occluded plant regions \cite{zhang2025wheat3dgs, stuart2025high, wang2025p3dfusion, chen2025plant,mcafee2025evaluation,ojo2024splanting}. Despite these advances, passive methods still face several challenges. They are often labor-intensive, requiring dense image capture and accurate image registration. Moreover, these reconstructions are typically performed offline and can be slow, limiting their applicability for large-scale phenotyping or real-time agricultural monitoring.

In contrast, active methods aim to autonomously guide data collection to improve reconstruction efficiency and completeness. A common approach in agricultural environments is next-best-view (NBV) planning \cite{burusa2022attention,zaenker2021viewpoint,menon2023nbv,cuaran2025active}, where a robot or imaging system actively selects the next camera viewpoint to maximize information gain. Most of of these approaches rely on occupancy grid representations, such as OctoMap \cite{hornung2013octomap}, to model known and unknown regions of the environment. OctoMap provides a probabilistic 3D voxel-based map that facilitates exploration and collision-free navigation. However, its main limitation lies in the raycasting process, which becomes computationally expensive as the map resolution increases. Building high-resolution OctoMaps for complex plant structures, where fine details such as leaves and stems matter, is therefore often impractical in real-world agricultural deployments.

To address these challenges, recent active mapping approaches have explored the use of Gaussian Splatting in domains outside agriculture, particularly in indoor environments \cite{tao2024rt, li2025activesplat}. Gaussian Splatting provides an explicit and differentiable scene representation that encodes both geometric and photometric information, enabling high-quality reconstruction and realistic novel-view rendering. However, a fundamental limitation of Gaussian Splatting for active mapping is that it does not explicitly represent unknown space, which is critical for exploration and viewpoint planning. Some works have addressed this limitation by using high-uncertainty Gaussians as proxies for frontiers to approximate unexplored regions \cite{tao2024rt}. However, this assumption is not always valid, as high-uncertainty Gaussians can also occur in already observed regions due to sparse observations, occlusions, or weak photometric constraints. Consequently, these approaches may incorrectly classify non-frontier regions as exploration targets.

Recent works have instead explored hybrid representations that combine occupancy-based maps with Gaussian Splatting, leveraging the complementary strengths of both representations \cite{jin2025activegs, jin2024gs, xu2025hgs}. In particular, a low-resolution OctoMap can provide explicit representations of free, occupied, and unknown space for exploration, frontier detection, and collision-free navigation and manipulation, while Gaussian Splatting captures fine-grained scene details with high fidelity. These hybrid approaches have been developed primarily for indoor environments, where scenes are relatively structured and depth and semantic observations are often more reliable. However, none of these methods has investigated target-aware mapping in horticultural environments, where severe occlusions, thin and deformable plant structures, and noisy depth and semantic measurements introduce additional challenges. In this work, we extend this hybrid representation to target-aware active mapping in horticultural environments and explicitly account for the noise characteristics of depth sensors and semantic segmentation models. 

Another challenge we address is the high memory consumption of 3DGS, particularly when reconstructing scenes that contain substantial background regions that are irrelevant to the target objects. Several recent works have therefore explored background pruning strategies that produce more compact Gaussian representations and cleaner plant reconstructions. Some approaches perform pruning as a post-processing step by projecting semantic masks onto the reconstructed scene to isolate specific plant structures, such as fruits or stems \cite{jiang2025cotton3dgaussians, zhang2025wheat3dgs, ojo2024splanting}. However, because these strategies are applied after training, the computational cost of optimizing irrelevant background Gaussians remains. Other approaches incorporate background removal before or during training by leveraging foundation models, such as Grounding DINO \cite{liu2024grounding} and Segment Anything \cite{kirillov2023segment}, to obtain semantic masks for pruning \cite{li2026object, wang2025p3dfusion, chen2025plant}. These methods, however, generally assume relatively accurate segmentation and have primarily been evaluated in controlled environments with isolated plants. In real-world horticultural environments, semantic segmentation is affected by illumination changes, occlusions, object scale, and visual similarity between plant structures and the background (see Fig. \ref{fig:sample_rgb_and_semantics} in Appendix \ref{sec:appendix-sample-viewpoints-with-semantics} with typical semantic noise found in real agricultural settings). Consequently, directly pruning Gaussians based on noisy segmentation can remove valid target geometry along with the background. This motivates the need for a pruning strategy that reduces the memory and computational cost of 3DGS while remaining robust to imperfect semantic perception.

As summarized in Table~\ref{tab:related_works}, to the best of our knowledge, ours is the first work to explore active semantic mapping using 3DGS in agricultural environments.

%% file: sections/methods.tex
% \vspace{-0.2cm}
\begin{figure*}[h]
\centerline{\includegraphics[width=170mm]{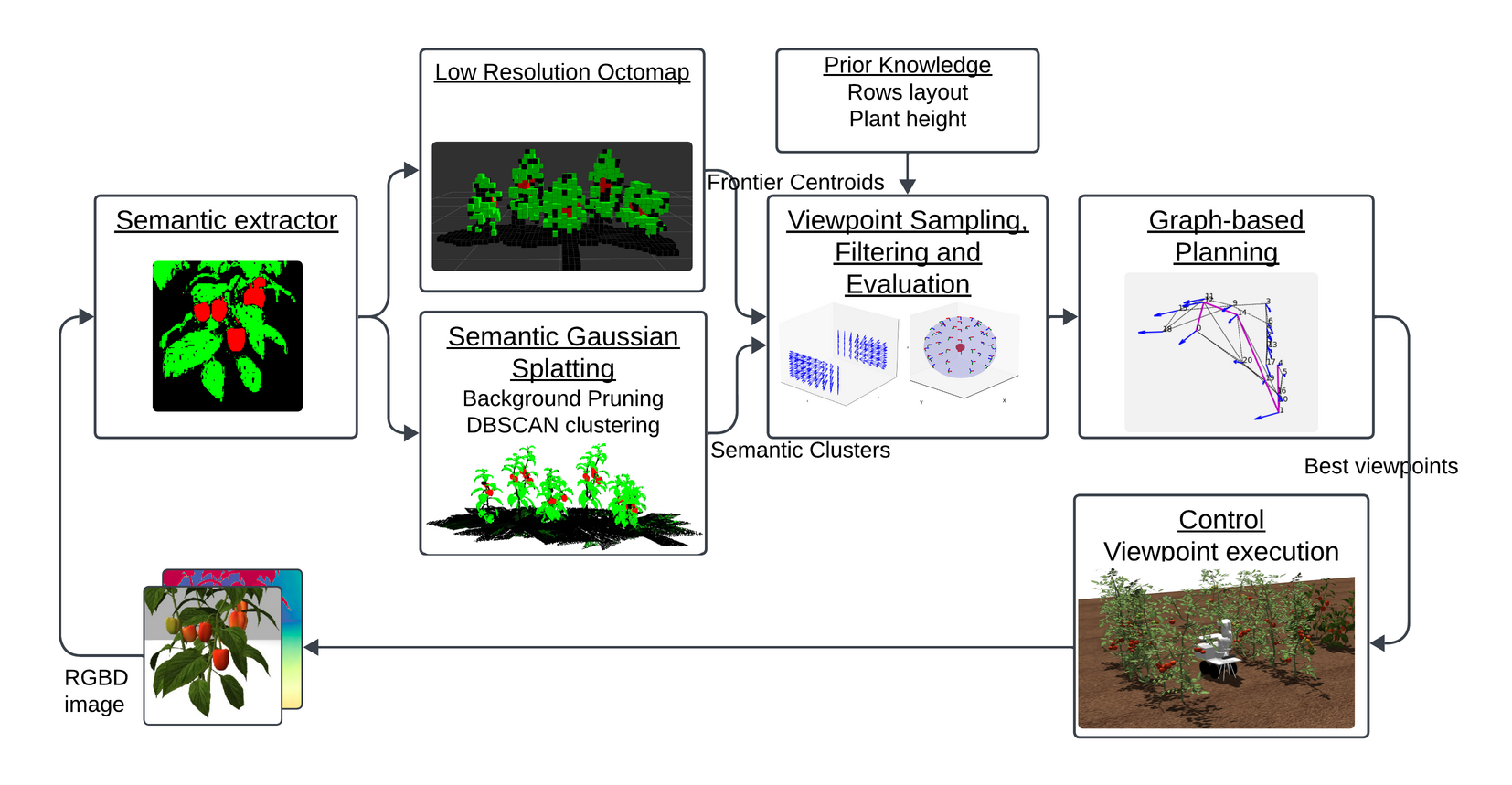}}
% \vspace{-0.2cm}
\caption{System overview. Our framework integrates two semantic representations: a low-resolution OctoMap for collision-free motion planning, frontier extraction, and viewpoint evaluation, and a 3DGS representation for high-fidelity reconstruction. Candidate exploitation viewpoints are sampled around semantic clusters, while exploration viewpoints are sampled around frontier centroids and along crop rows. A graph-based planner determines the optimal sequence of viewpoints by considering information gain and actuation cost.}
\vspace{-0.1cm}
\label{fig:system_overview}
% \vspace{-14pt}
\end{figure*}   

We aim to build a semantic 3D reconstruction of plants from RGBD observations, focusing on a specific target (e.g., fruits). Fig. \ref{fig:system_overview} presents an overview of our system. The system receives RGBD images from the manipulator's tip camera, and outputs a sequence of viewpoints that progressively improve the reconstruction of the target semantics. At each observation, a semantic extractor predicts pixel-level class labels. These semantic images and depth data are used to update two complementary representations: a low-resolution semantic OctoMap to maintain occupancy information, and a dense semantic 3DGS model for detailed reconstruction. The gaussian representation enables identifying target regions by clustering semantic gaussians. A viewpoint sampling, filtering and evaluation module takes these semantic clusters, sample multiple viewpoints and evaluate them based on the occupancy information. Finally, a graph-based planner selects and executes a subset of viewpoints that balance information gain and actuation cost. The following subsections describe each module in detail.

\subsection{Semantic extractor}
The semantic extractor takes an RGB image as input and produces a per-pixel semantic label and segmentation confidence. We consider the semantic classes $\mathcal{S}=\{$ \textit{fruits, leaves, background} $\}$. The corresponding semantic masks are used to update both the low-resolution OctoMap and the semantic 3DGS model, ensuring consistency between the coarse and dense representations.

\subsection{Low-resolution OctoMap}
OctoMap \cite{hornung2013octomap} has been widely used for active mapping and exploration. In our system, we utilize it primarily for viewpoint planning rather than detailed mapping. This allows us to maintain a low-resolution OctoMap, which is computationally efficient and sufficient for our planning needs. We use a semantic extension \cite{asgharivaskasi2023semantic} that takes RGBD observations and semantic labels as input and updates the occupancy and class probability of the environment. Specifically, Semantic Octomap creates a probabilistic semantic octree where each voxel $x$ encodes a categorical distribution $p_c(x)$ over different classes as well as a binary occupancy probability value $po(x)$. This semantic OctoMap enables the computation of a semantics-aware information gain for viewpoint evaluation, as in \cite{cuaran2025active}, and collision checking during planning.

\subsection{Semantic Gaussian Splatting}
We represent the environment as a collection of 3D gaussians, each parameterized by position $\boldsymbol{\mu} \in \mathbb{R}^3$, radius $r \in \mathbb{R}$, color $\mathbf{c} \in \{R,G,B\}$, opacity $o \in [0,1]$, and semantic label $\mathbf{s} \in \mathcal{S}$. These parameters are optimized by minimizing the photometric, depth, and semantic discrepancies between rendered and input RGBD images. Our approach builds upon SGS-SLAM~\cite{li2024sgs}, a work focused on 3DGS-based Semantic SLAM. Similar to this work, we use isotropic 3D Gaussians defined as follows:
\begin{equation}
\scalebox{0.95}{$
    f(\mathbf{x}) = o \exp\left( -\frac{\lVert \mathbf{x} - \boldsymbol{\mu} \rVert^2}{2r^2} \right)$
}
\label{eq:f_x}
\end{equation}

where $\mathbf{x} \in \mathbb{R}^3$ is a 3D point in space.

Color images $\mathbf{C(\mathbf{p})}$ are rendered by alpha-compositing the 2D projections of all gaussians onto the image plane, as follows:
\begin{equation}
\scalebox{0.95}{$
    C(\mathbf{p}) = \sum_{i=1}^{N} \mathbf{c_i} f_{2D_i}(\mathbf{p}) \prod_{j=1}^{i-1} (1 - f_{2D_j}(\mathbf{p}))
    $}
\label{eq:C_p}
\end{equation}

where $\mathbf{c_i}$ is the color value of the $i$th gaussian, $f_{2D_j}(\mathbf{p})$ is the value of $f$ projected onto the 2D image plane at the pixel $\mathbf{p}=(u,v)$, leveraging the camera intrinsic matrix and the camera pose. Semantic, depth and silhouette (visibility) images are rendered analogously by replacing $\mathbf{c_i}$ in equation \ref{eq:C_p} by $\mathbf{s_i}$, depth value $\mathbf{z_i}$ and $1.0$ respectively.

% Unlike \cite{keetha2024splatam} and \cite{li2024sgs}, we do not use 3DGS for localization which requires continuous camera tracking, but instead we leverage 3DGS to refine initial camera poses provided by visual odometry and the manipulator forward kinematics. In addition, the following implementation choices were crucial to make our approach efficient and robust to sensor noise.

We assume in this work that accurate camera pose estimation is available, which, to a certain extent, is ensured by the manipulator's forward kinematics. Therefore, unlike \cite{li2024sgs}, we do not use 3DGS for localization, as it requires continuous camera tracking and thus introduces additional computational overhead.

The initialization, mapping and background pruning strategies described below are designed to make our approach efficient and robust to depth and semantic segmentation noise.

\textbf{Initialization.} For each new RGB-D observation, Gaussian positions are initialized from the corresponding depth image as in the original implementation. We initialize one Gaussian per valid pixel.  Although these noisy values may differ from the true scene geometry, the Gaussian positions are subsequently refined during optimization using depth, color, and semantic information. We only consider depth measurements below a maximum depth threshold, $max\_range$, to avoid introducing Gaussians corresponding to distant background regions, such as adjacent crop rows.
Additionally, the semantic features of newly initialized Gaussians are assigned neutral values, allowing them to gradually converge to the correct semantic labels through multiple noisy observations. This initialization, together with semantic supervision, enables the background pruning strategy described below.

% However, real-world depth measurements are often incomplete and contain invalid or missing (zero) values. To reduce missing surfaces in the reconstruction, we fill missing depth values within confident foreground regions using nearest-neighbor interpolation based on the Euclidean distance transform. This approach assigns each invalid pixel the depth value of its closest valid neighbor.

% \textbf{Tracking.} For each new observation, we first refine the camera pose by freezing the gaussian parameters and optimizing the camera pose, running some tracking iterations. This helps refine the initial camera pose coming from visual odometry and forward kinematics. We do it considering depth, color and semantic losses as in \cite{li2024sgs}.

% \begin{equation}
% \scalebox{0.95}{$
%     L_t = \sum_{\mathbf{p}} \left( Sil(\mathbf{p}) > T_s \right) 
% \left( \lambda_1 L_d(\mathbf{p}) + \lambda_2 L_c(\mathbf{p}) +\lambda_3 L_s(\mathbf{p}) \right)$}
% \label{eq:L_t}
% \end{equation}

% With:
% \[
% L_d(\mathbf{p}) =|D(\mathbf{p})-D_{gt}(\mathbf{p})|
% \]
% \[
% L_c(\mathbf{p}) =|C(\mathbf{p})-C_{gt}(\mathbf{p})|
% \]
% \[
% L_s(\mathbf{p}) =|conf^2(\mathbf{p}) \left (S(\mathbf{p})-S_{gt}(\mathbf{p}) \right )|
% \]

%%%%%%
\textbf{Robust Mapping.} Similar to \cite{li2024sgs}, we jointly optimize geometric, photometric, and semantic consistency using the following mapping loss:

\begin{equation}
\label{eq:L_m}
\scalebox{0.92}{$
L_m =
\sum_{\mathbf{p}\in F}
\left(
\lambda_1 L_d(\mathbf{p})
+
\lambda_2 L_c(\mathbf{p})
\right)
+
\lambda_3
\sum_{\mathbf{p}}
L_s(\mathbf{p})
$}
\end{equation}

where

\begin{equation*}
\scalebox{0.90}{$
L_d(\mathbf{p}) =
\left|
D(\mathbf{p})-
D_{gt}(\mathbf{p})
\right|
$}
\end{equation*}
\vspace{-5mm}

\begin{equation*}
\scalebox{0.88}{$
L_c(\mathbf{p}) =
\alpha
\left|
C(\mathbf{p})-
C_{gt}(\mathbf{p})
\right|
+
(1-\alpha)
\left(
1-
SSIM\!\left(C(\mathbf{p}),C_{gt}(\mathbf{p})\right)
\right)
$}
\end{equation*}
\vspace{-5mm}

\begin{equation}
\label{eq:segmentation_loss}
\scalebox{0.92}{$
L_s(\mathbf{p}) =
\left|
\mathrm{conf}(\mathbf{p})^2
\left(
S(\mathbf{p})
-
S_{gt}(\mathbf{p})
\right)
\right|
$}
\end{equation}

where $\lambda_1$, $\lambda_2$, $\lambda_3$, and $\alpha$ are weighting hyperparameters summarized in Table~\ref{tab:parameters}, and $C_{gt}$, $D_{gt}$, and $S_{gt}$ denote the input color, depth, and semantic images, respectively.

Unlike previous works \cite{li2024sgs}, we weight the semantic loss by the squared confidence of the semantic extractor, $\mathrm{conf}(\mathbf{p})^2$, thereby reducing the influence of uncertain semantic predictions. Furthermore, the depth and color losses are computed only over foreground pixels, preventing the background from influencing the optimization of foreground Gaussians, particularly those near object boundaries. The foreground mask $F$ is defined as:

\begin{equation}
\label{eq:foreground_mask}
F =
\left\{
\mathbf{p}
\;\middle|\;
S(\mathbf{p}) \neq \mathrm{background}
\;\land\;
\mathrm{conf}(\mathbf{p}) > \tau
\right\},
\end{equation}

where $\tau$ is an empirically selected confidence threshold. Unlike the depth and color losses, the semantic loss is evaluated over the entire image. This encourages Gaussians corresponding to background regions to converge toward the background class, enabling their subsequent removal through the proposed semantic pruning strategy.

\textbf{Background Pruning.} In addition to the standard pruning operations used in 3DGS, we introduce a semantic pruning strategy to remove Gaussians that represent background regions. Unlike post-processing approaches that remove background Gaussians after training, our strategy operates during mapping, allowing the representation to remain compact throughout optimization and reducing the computational cost of subsequent mapping iterations.

At each pruning iteration, we evaluate the semantic representation of each Gaussian and remove those whose $\ell_2$ distance to the background semantic label falls below a predefined threshold. This strategy is enabled by the semantic initialization and loss design described above. In particular, newly initialized Gaussians are assigned neutral semantic features, while semantic supervision is applied over the entire image, including background regions. Consequently, Gaussians initialized from background or spurious depth measurements can gradually acquire background semantics through successive observations and are subsequently removed during pruning. This is particularly beneficial under noisy or incomplete depth measurements, where direct initialization can introduce Gaussians at incorrect locations or in regions that should not be reconstructed.

By removing background Gaussians during mapping rather than after training, our strategy maintains a more compact representation throughout optimization, thereby reducing both GPU memory consumption and Gaussian Splatting mapping time while preserving the target plant geometry.

It is worth noting that, although our implementation is built upon the SGS-SLAM framework~\cite{li2024sgs}, the proposed method is largely framework-agnostic and can be integrated into other Gaussian Splatting pipelines with only minor modifications. To demonstrate this generality, we also adapted our method to two additional Gaussian Splatting frameworks, 2DGS~\cite{huang20242d} and GS3LAM~\cite{li2024gs3lam}, as part of our experimental evaluation.

\subsection{Viewpoint Sampling, Filtering, and Evaluation}
To identify informative viewpoints for active mapping, we exploit the complementary strengths of our two map representations. The 3DGS model is used to identify reliable semantic targets (\textit{fruits}) and generate exploitation viewpoints, whereas the low-resolution OctoMap is used to generate exploration viewpoints.

To generate exploitation viewpoints, we apply DBSCAN clustering to Gaussians whose semantic features have converged to the target class (\textit{fruit}). Candidate viewpoints are then uniformly sampled on a sphere around the centroid of each fruit cluster. Exploration viewpoints are generated similarly by uniformly sampling viewpoints around frontier centroids. To this end, frontier voxels are identified in the OctoMap as free voxels adjacent to both occupied and unknown voxels within a 26-connected neighborhood. These frontier voxels are subsequently grouped using voxel-grid clustering, and each cluster is represented by its centroid, yielding a compact set of frontier centroids.

To bootstrap the mapping process, we additionally define a set of exploration viewpoints on planes parallel to the crop rows using prior knowledge of the row spacing and average plant height. These viewpoints are particularly useful at the beginning of the mapping process, before either the OctoMap or the 3DGS model has been initialized.

Because many candidate viewpoints may lie outside the manipulator's workspace, we project them onto the nearest reachable pose. To this end, we discretize the manipulator's joint space and compute forward kinematics to obtain a set of feasible 6-DoF end-effector poses. For each candidate viewpoint, the nearest reachable pose is determined using a nearest-neighbor search. This procedure ensures that all selected viewpoints are kinematically feasible.

Finally, the occupancy map is used to evaluate the information gain of each candidate viewpoint. Specifically, exploration viewpoints are evaluated using the Unknown Voxel Count (UVC) metric \cite{delmerico2018comparison}, whereas exploitation viewpoints are evaluated using the Occlusion- and Semantic-Aware Multi-Class Entropy with Proximity Count (OSAMCEP) metric \cite{cuaran2025active}. The information gain values are normalized independently for each viewpoint category, and the top $K$ viewpoints are selected to construct the planning graph.

\subsection{Graph-based planning}
Inspired by \cite{zaenker2023graph}, once we have a set of viewpoints with their corresponding information gains, we insert each viewpoint and its associated joint configuration as a node in a graph. We connect each node to $N_{near}$ neighboring nodes based on proximity in the manipulator's joint space. In this way, we encourage the manipulator to perform smooth and efficient transitions between viewpoints.

We use a best-first path search algorithm to find the next sequence of viewpoints to be executed. The algorithm employs a priority queue (initialized with the current camera pose) to store and retrieve nodes based on their utility values. At each step, the node with the highest utility is extracted from the priority queue and expanded. When unexpanded neighbors are discovered, their utility is updated as the sum of their predecessor’s utility and their own, and they are inserted into the priority queue. This process continues until no nodes remain in the queue.
Finally, the optimal path is obtained by backtracking from the highest-utility node to the starting position. We execute the top $K_{exec}$ viewpoints along this best path before replanning.

\subsection{Control}
We use the ROS MoveIt package to execute manipulation viewpoints following collision-free trajectories.

\subsection{Parallel Mapping}
After a new viewpoint observation becomes available, the 3DGS module initializes new Gaussians and performs $K_{\text{mapping}}$ optimization iterations. At each iteration, a previously captured frame is randomly selected to perform a mapping optimization step. In parallel, the low-resolution OctoMap is updated with the new observation. Once both mapping processes are complete, the updated OctoMap is used to sample and evaluate new candidate viewpoints, while the updated 3DGS model is used to compute new semantic clusters.

% Unlike the original SLAM implementation \cite{li2024sgs}, our method does not require a keyframe selection strategy, since each new viewpoint is selected based on the expected information gain from the current map.

%% file: sections/experiments.tex
\begin{table}[htbp]
\centering
\caption{Main hyperparameters used during evaluation. All the hyperparameters are available in the supplementary material.}
\vspace{-0.2cm}
\resizebox{0.5\textwidth}{!}{
\begin{tabular}{c|c|c|l}
    \toprule
    \textbf{Category} & \textbf{Parameter} & \textbf{Value} & \textbf{Description} \\
    \midrule
    \textbf{Low resolution} & $\delta S$ [m] & 0.05 & Map resolution \\
    \textbf{Octomap}& $max\_range$ [m] & 1.0 & Max depth range for mapping \\
    \midrule
    \textbf{High resolution} & $\delta S$ [m] & 0.01 & Map resolution \\
    \textbf{Octomap}& $max\_range$ [m] & 1.0 & Max depth range for mapping \\
    \midrule
    \textbf{Segmentation} & $P_{gt}$ & 0.7 & Probability of correct classification \\
    {} & {} & {} & during semantic segmentation \\
    \midrule
    \textbf{Viewpoint} & $r$ [m] & 0.4 & Radius of sphere for viewpoint sampling \\
    \textbf{Sampling} & $N_{\phi}$ & 10 & Number of azimuth samples \\
    & $N_{\theta}$ & 5 & Number of elevation samples \\
    \midrule
    \textbf{Graph-based} & $K$ & 20 & Number of viewpoints for graph \\
    \textbf{planning} & $N_{near}$ & 4 & Number of neighbors per node\\
    & $K_{exec}$ & 4 & Viewpoints executed before replanning\\
    \midrule
    \textbf{DBSCAN} & $\epsilon$ [m] & 0.02 & Radius of a neighborhood \\
     & $min-samples$ [m] & 10 & Minimum points per cluster \\
    \midrule
    \textbf{Gaussian} & $\lambda_1$ & 0.5 & Weight for depth loss \\
    \textbf{Splatting}
    & $\lambda_2$ & 1.0 & Weight for rgb loss \\
    (SGS-SLAM, GS3LAM and 2DGS)& $\lambda_3$ & 0.1 & Weight for semantic loss \\
    & $\alpha$ & 0.8 & Second weight for rgb loss \\
    & $\tau$ & 0.4 & Confidence threshold \\
    & $K_{mapping}$ & 60 & Number of optimization steps \\
    
    \bottomrule 
\end{tabular}
}
\label{tab:parameters}
% \vspace{-0.6cm}
\end{table}

We validate our approach through simulation, controlled laboratory experiments, and real greenhouse experiments. These corresponding environments can be seen in Fig. \ref{fig:merged_environments}. Simulation enables a quantitative evaluation of reconstruction accuracy by providing high-quality ground-truth data. Controlled laboratory experiments allow us to assess the proposed method under static conditions, without the influence of environmental disturbances. Finally, real greenhouse experiments demonstrate the applicability of our approach in realistic agricultural environments, where challenges such as wind-induced plant motion and illumination variations can significantly affect perception and reconstruction. 

\textbf{Simulation environment.} We construct a realistic horticultural environment in the Gazebo simulator. The simulated scene comprises six crop rows, including two variants of bell pepper plants and one variant of tomato plants. These plant models \cite{cuaran2026visual} exhibit variations in appearance, height, cluster size, and occlusion characteristics. To increase environmental diversity, we generate five replicas of this setup by randomizing the orientation of the plant models, obtaining 30 different rows in total.

To evaluate the robustness of our framework to imperfect semantic perception, we simulate segmentation errors by assigning each segmented object mask its correct semantic label with probability $P$, and an incorrect label with probability $1-P$. When an incorrect label is assigned, it is selected uniformly from the remaining semantic classes. This object-level noise model emulates classification errors commonly produced by semantic segmentation systems while preserving the spatial consistency of individual object masks. To maintain consistency with this noise model, we generate semantic confidence maps by assigning a confidence of $P$ to the predicted class when the prediction is correct and $\frac{1-P}{2}$ when it is incorrect.

In addition, we use SimSense\footnote{\href{https://github.com/angli66/simsense}{https://github.com/angli66/simsense}}, a GPU-accelerated depth sensor simulator based on stereo matching, to generate noisy depth maps that resemble those produced by real RGB-D sensors.

\textbf{Laboratory environment.} Our controlled environment consists of four synthetic plants, each separated by 0.5 m, containing bell peppers and lime fruits. We generate 10 different scenes by randomly varying the orientation of the plants.

\textbf{Real greenhouse environment.} We use eight potted bell pepper plants arranged in two rows of four plants each. The plants are spaced 0.5 m apart and bear green, red, and yellow bell peppers. We generate 10 distinct scenes by randomly varying the orientation of the plants.

\textbf{Robot setup.} Our mobile manipulator consists of a 6-DoF XArm Lite6 manipulator mounted on a Terrasentia ground robot. An RGB-D camera with a resolution of (640x480) pixels is attached to the manipulator's end effector, with its optical axis aligned with the rotation axis of the wrist joint. In simulation, we use ground-truth camera poses. For real-world experiments, the tip camera pose is estimated by combining the built-in SLAM system of a ZED2 camera mounted on the front of the robot with the manipulator's forward kinematics.

For semantic perception, we use color-based segmentation in simulation. In laboratory and field experiments, this module is replaced by Detic \cite{zhou2022detecting}, an open-vocabulary detector used for fruit detection and segmentation, and CrossGSD-Seg \cite{gao2024bridging}, which is used for vegetation segmentation. Both models return object masks and confidence scores.

The main hyperparameters used in our experiments are summarized in Table~\ref{tab:parameters}. Additional implementation details are provided in the supplementary material.

All simulations and real-world experiments are conducted on an NVIDIA Jetson AGX Orin platform equipped with 32 GB of unified memory.

\textbf{Experimental procedure.} For each crop row consisting of four or five plants, we define a fixed sequence of six waypoints (three on each side of the row) that the mobile manipulator follows throughout all trials, thereby minimizing the influence of navigation variability. The robot performs mapping along the row by traversing one side, turning at the end of the row, and returning along the opposite side. At each waypoint, it selects and executes up to 12 successful viewpoints for data acquisition. We consider three semantic classes: fruits, leaves, and background, with fruits designated as the target objects for mapping.

\textbf{Baselines.} We compare our framework against two classes of baselines. First, we evaluate the complete hybrid active mapping framework against a Semantic OctoMap-based active mapping approach. Second, we evaluate the semantic Gaussian Splatting module within our framework against state-of-the-art semantic Gaussian Splatting methods, including SGS-SLAM \cite{li2024sgs}, GS3LAM \cite{li2024gs3lam}, and 2DGS \cite{huang20242d}.

\textit{Semantic OctoMap}. A purely OctoMap-based active mapping approach, which remains the predominant map representation in horticultural robotics~\cite{cuaran2025active, burusa2022attention, zaenker2023graph}. Within our modular framework, this baseline is implemented by removing the 3D Gaussian Splatting module while retaining the semantic OctoMap with a voxel resolution of 0.01~m, which is five times finer than the resolution used in our hybrid approach. The resulting OctoMap is used to identify fruit clusters and frontier centroids, and evaluate candidate viewpoints, while all remaining modules are kept unchanged to ensure a fair comparison.

\textit{SGS-SLAM}. A semantic 3D Gaussian Splatting framework that jointly optimizes the 3D scene representation and camera poses. Since our viewpoints are sparsely sampled and therefore unsuitable for continuous SLAM, we disable the tracking module and optimize only the mapping component. The optimization objective includes RGB, depth, and semantic losses.

\textit{GS3LAM}. Another semantic SLAM framework based on 3D Gaussian Splatting. Unlike SGS-SLAM, which employs an $\ell_1$ semantic loss, GS3LAM uses a multi-class cross-entropy semantic loss and jointly trains a semantic classifier. Its objective function also includes Gaussian scale regularization in addition to the RGB, depth, and semantic losses. As with SGS-SLAM, the tracking module is disabled.

\textit{2DGS}. An adaptation of Gaussian Splatting designed for accurate surface reconstruction. Unlike 3DGS, which represents the scene using volumetric Gaussians, 2DGS models surfaces as planar disks (surfels), enabling direct surface optimization. The original implementation optimizes RGB reconstruction, depth distortion, and normal consistency losses. For a fair comparison, we extend it by incorporating depth and semantic losses.

Unlike the OctoMap baseline and our approach, two of our Gaussian Splatting baselines are evaluated offline. For each scene, we first run our hybrid active mapping framework, using SGS-SLAM (ours) as backbone, and record the RGB, depth, semantic, and confidence images together with the corresponding camera poses. The remaining Gaussian Splatting baselines (2DGS and GS3LAM) are then evaluated using the recorded data. This evaluation protocol significantly reduces the computational cost of the experiments by eliminating the need to repeat data acquisition for each baseline, while ensuring that all methods are evaluated using the same observations and camera trajectories for a fair comparison.

\textbf{Metrics.}
We evaluate the proposed approach and the baselines in terms of both fruit 3D reconstruction accuracy and computational efficiency. To assess reconstruction accuracy, we use standard geometric metrics, including the Chamfer Distance (CD), Precision ($P$), Recall ($R$), and F1 score ($F_1$). A distance threshold of 0.01 m is used to compute Precision and Recall, consistent with the resolution of the Semantic OctoMap baseline.

We also evaluate novel-view synthesis (NVS) quality using the Peak Signal-to-Noise Ratio (PSNR), Structural Similarity Index Measure (SSIM), Mean Absolute Depth Error (Depth L1), and Mean Intersection over Union (mIoU). Because our primary objective is accurate fruit reconstruction, PSNR, SSIM, and Depth L1 are computed only over pixels corresponding to fruit regions, with all other semantic classes masked out. Ground-truth images for NVS evaluation are obtained by sampling random viewpoints directed toward fruit clusters during the same mapping episodes. These viewpoints are used exclusively for evaluation and are not incorporated into either the OctoMap or Gaussian Splatting reconstruction. This protocol ensures that the evaluation images are captured from novel viewpoints while maintaining camera poses that are consistent with the current world coordinate frame.

For the laboratory and real greenhouse experiments, we use the semantic predictions generated by Detic and CrossGSD-Seg, retaining only predictions with confidence values above a threshold $\tau$, as an approximation of the ground-truth semantics. Similarly, camera depth measurements with zero values masked out are used as an approximation of the ground-truth depth.

Finally, we report the average runtime per viewpoint, the number of Gaussians, and peak GPU memory usage as measures of computational efficiency.

% The \textit{Chamfer Distance} quantifies the geometric discrepancy between the reconstructed fruit point cloud $\mathcal{P}$ and the ground-truth fruit point cloud $\mathcal{Q}$ as
% \begin{equation*}
% \text{CD}(\mathcal{P}, \mathcal{Q}) = \frac{1}{|\mathcal{P}|} \sum_{p \in \mathcal{P}} \min_{q \in \mathcal{Q}} |p - q|_2 +
% \frac{1}{|\mathcal{Q}|} \sum_{q \in \mathcal{Q}} \min_{p \in \mathcal{P}} |q - p|_2.
% \end{equation*}

% To evaluate point-level correspondence accuracy, we define \textit{Precision} and \textit{Recall} as
% \begin{equation}
% P = \frac{|{p \in \mathcal{P} : d(p, \mathcal{Q}) < \tau }|}{|\mathcal{P}|}, \quad
% R = \frac{|{q \in \mathcal{Q} : d(q, \mathcal{P}) < \tau }|}{|\mathcal{Q}|},
% \end{equation}
% where $d(x, \mathcal{Y})$ denotes the minimum Euclidean distance from point $x$ to the set $\mathcal{Y}$, and $\tau$ is a fixed distance threshold. We set $\tau$ at 0.015 m, the coarsest octomap resolution considered in this study. The harmonic mean of Precision and Recall yields the \textit{F1 Score}:
% \begin{equation}
% F_1 = 2 \times \frac{P \times R}{P + R}.
% \end{equation}

% In practice, Recall represents fruit coverage, and it is a good indicator of our method's ability to find viewpoints that reveal fruit areas despite the self-occlusion characteristics of plants.

%% file: sections/results.tex
\begin{figure*}[htbp]
\centerline{\includegraphics[width=180mm]{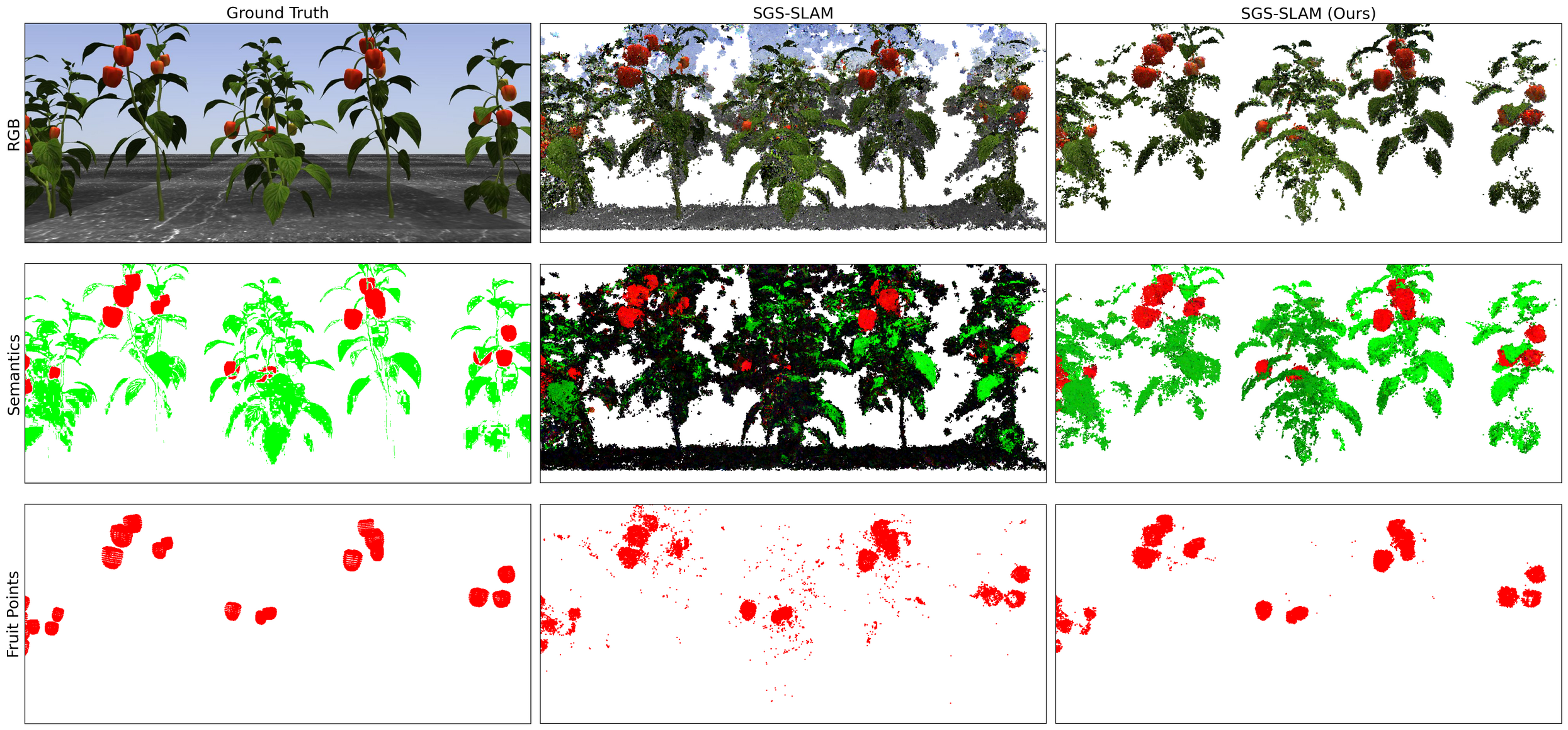}}
\vspace{-0.2cm}
\caption{Sample reconstructions of a bell pepper row from our simulated environment. Compared with the SGS-SLAM baseline, our 3DGS-based approach produces cleaner and less noisy RGB, semantic, and fruit point clouds. Additional examples are provided in Appendix~\ref{sec:appendix-sample-reconstructions}.}
\vspace{-0.1cm}
\label{fig:sgs_simulation_outputs_row1}
\vspace{-14pt}
\end{figure*} 

In the results reported in this section, we denote by \textit{Backbone (Ours)} (e.g., SGS-SLAM (Ours)) the corresponding baseline augmented with our robust mapping and background pruning strategies (Section~\ref{sec:methods}), in contrast with its original (unmodified) formulation.

\begin{figure*}[htbp]
\centerline{\includegraphics[width=160mm, height=200mm]{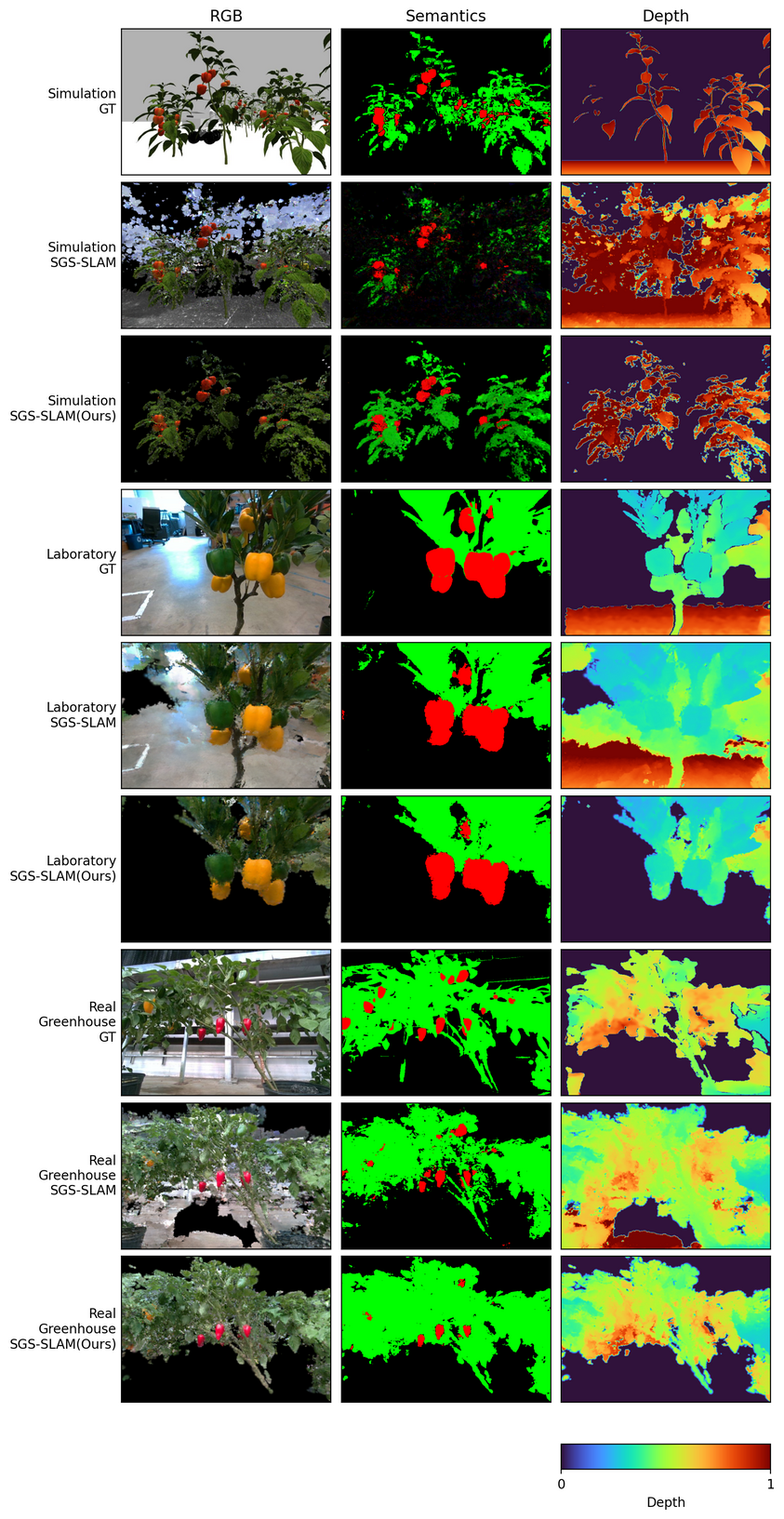}}
\vspace{-0.2cm}
\caption{Novel views generated using SGS-SLAM and our approach in different environments.  We refer to the depth and semantic images obtained from the camera sensors and segmentation model, respectively, as ground truth (GT). Note that these reference measurements may contain errors and are therefore not necessarily perfect in the laboratory and real greenhouse environments. Additional examples are provided in Appendix \ref{sec:appendix-nvs} for other GS methods. }
\vspace{-0.1cm}
\label{fig:nvs-sgs-grid}
\vspace{-14pt}
\end{figure*}

\textbf{Fruit Reconstruction Accuracy.}
Table~\ref{table:reconstruction-metrics-simulation} reports fruit-level reconstruction accuracy in simulation under the most challenging condition considered in this work, namely, simultaneous depth and segmentation noise. Across all three Gaussian Splatting backbones, augmenting the vanilla baselines with our robust mapping and background pruning strategies substantially improves reconstruction accuracy, reducing Chamfer Distance by 4--7$\times$ and more than doubling Precision for GS3LAM and 2DGS. In contrast, Recall remains comparatively stable across most backbones, indicating that the vanilla baselines already recover most fruit surfaces but represent them with a large number of spurious, noisy points. SGS-SLAM (Ours) achieves the best performance across all four metrics and is therefore adopted as the representative configuration of our framework in the remaining sections. One exception is 2DGS (Ours), whose Recall decreases from 0.927 to 0.822 relative to its vanilla counterpart despite a substantial gain in Precision. This suggests that the proposed strategies remove a larger fraction of valid fruit surfels along with spurious ones, potentially due to suboptimal hyperparameter tuning for this particular backbone.

Compared with the Semantic OctoMap baseline, all Gaussian Splatting methods, including their unmodified baselines, achieve a higher F1 score, highlighting a fundamental limitation of occupancy-grid representations under noisy conditions. In particular, Semantic OctoMap attains a comparatively high Recall (0.740) but the lowest Precision of all methods (0.346), as noisy depth and segmentation measurements corrupt individual voxels that cannot be corrected through joint photometric, geometric, and semantic optimization, unlike our 3DGS-based representation. Overall, our best configuration, SGS-SLAM (Ours), nearly doubles the F1 score of the Semantic OctoMap baseline.

Fig.~\ref{fig:sgs_simulation_outputs_row1} illustrates this improvement qualitatively for one simulated crop row. Compared with the vanilla SGS-SLAM baseline, our approach produces markedly cleaner RGB, semantic, and fruit point clouds, with visibly fewer spurious points scattered around the fruit clusters, consistent with the precision gains reported in Table~\ref{table:reconstruction-metrics-simulation}.

\begin{table*}[htbp]
\centering
\caption{Reconstruction performance in the simulated environment under noisy conditions, averaged over 30 scenes.}
{
\resizebox{0.6\textwidth}{!}{
\begin{tabular}{c c c c c}
    \toprule
    \textbf{Method} & \textbf{Chamfer Distance$\downarrow$ [m]}& \textbf{Precision$\uparrow$} & \textbf{Recall$\uparrow$} & \textbf{F1 Score$\uparrow$}\\
    \midrule
    \textbf{Semantic Octomap} & 0.101$\pm$0.025  & 0.346$\pm$0.127 & 0.740$\pm$0.214 & 0.466$\pm$0.155\\
    \textbf{GS3LAM}& 0.124$\pm$0.035  & 0.426$\pm$0.118 & 0.929$\pm$0.046 & 0.574$\pm$0.111\\
    \textbf{2DGS} & 0.096$\pm$0.024  & 0.434$\pm$0.123 & 0.927$\pm$0.041 & 0.594$\pm$0.119\\
    \textbf{SGS-SLAM} & 0.076$\pm$0.028  & 0.783$\pm$0.092 & 0.778$\pm$0.053 & 0.777$\pm$0.063\\
    \textbf{GS3LAM (Ours)}& 0.024$\pm$0.011  & 0.906$\pm$0.042 & \second{0.933$\pm$0.030} & \second{0.919$\pm$0.029}\\
    \textbf{2DGS (Ours)} & \second{0.023$\pm$0.011}  & \second{0.907$\pm$0.054} & 0.822$\pm$0.089 & 0.861$\pm$0.071\\
    \textbf{SGS-SLAM (Ours)}& \best{0.013$\pm$0.002}  & \best{0.936$\pm$0.019} & \best{0.958$\pm$0.026} & \best{0.946$\pm$0.016}\\
    \bottomrule
\end{tabular}
}
}
\label{table:reconstruction-metrics-simulation}
\vspace{-3mm}
\end{table*}

\begin{table*}[htbp]
\centering
\caption{Novel View Synthesis Performance Across Environments under noisy conditions}
{
\resizebox{0.8\textwidth}{!}{
\begin{tabular}{c c c c c c}
    \toprule
    \textbf{Environment} & \textbf{Method} & \textbf{PSNR[dB]$\uparrow$}& \textbf{SSIM$\uparrow$} & \textbf{Depth L1[m]$\downarrow$} & \textbf{MIOU$\uparrow$}\\
    \midrule
    \textbf{} & \textbf{GS3LAM} & 11.655$\pm$0.905  & 0.136$\pm$0.024 & 0.079$\pm$0.013 & 0.289$\pm$0.051\\
    \textbf{} & \textbf{2DGS} & 13.547$\pm$0.839  & 0.186$\pm$0.027 & 0.076$\pm$0.015 & 0.298$\pm$0.031\\
    \textbf{Simulation} & \textbf{SGS-SLAM} & 12.258$\pm$0.847  & 0.189$\pm$0.024 & 0.105$\pm$0.017 & 0.294$\pm$0.045\\
    \textbf{(30 scenes)} & \textbf{GS3LAM (Ours)}& \best{16.559$\pm$1.231}  & \second{0.222$\pm$0.040} & 0.116$\pm$0.021 & \second{0.692$\pm$0.030}\\
    \textbf{} & \textbf{2DGS (Ours)} & \second{16.348$\pm$1.018}  & 0.219$\pm$0.038 & \second{0.070$\pm$0.024} & 0.642$\pm$0.057\\
    \textbf{} & \textbf{SGS-SLAM (Ours)}& 16.156$\pm$0.993  & \best{0.238$\pm$0.032} & \best{0.052$\pm$0.012} & \best{0.713$\pm$0.049}\\
    \midrule
    \textbf{} & \textbf{GS3LAM}& 12.992$\pm$0.454  & 0.239$\pm$0.018 & 0.045$\pm$0.010 & 0.511$\pm$0.056\\
    \textbf{} & \textbf{2DGS}& 14.503$\pm$0.867  & 0.308$\pm$0.034 & 0.045$\pm$0.014 & 0.570$\pm$0.050\\
    \textbf{Laboratory} & \textbf{SGS-SLAM} & 14.174$\pm$0.667  & \second{0.322$\pm$0.023} & \second{0.041$\pm$0.012} & 0.599$\pm$0.030\\
    \textbf{(10 scenes)} & \textbf{GS3LAM (Ours)}& 14.160$\pm$0.362  & 0.254$\pm$0.017 & 0.073$\pm$0.009 & 0.691$\pm$0.025\\
    \textbf{} & \textbf{2DGS (Ours)} & \best{15.055$\pm$0.534}  & 0.303$\pm$0.024 & \best{0.038$\pm$0.009} & \second{0.691$\pm$0.024}\\
    \textbf{} & \textbf{SGS-SLAM (Ours)} &  \second{15.002$\pm$0.482}  & \best{0.333$\pm$0.016} & 0.049$\pm$0.008 & \best{0.692$\pm$0.017}\\
    \midrule
    \textbf{} & \textbf{GS3LAM}& 11.479$\pm$0.529  & 0.130$\pm$0.007 & 0.103$\pm$0.022 & 0.377$\pm$0.020\\
    \textbf{} & \textbf{2DGS}& 12.884$\pm$0.754  & \second{0.183$\pm$0.021} & 0.090$\pm$0.016 & 0.400$\pm$0.025\\
    \textbf{Real Greenhouse} & \textbf{SGS-SLAM} & 12.309$\pm$0.939  & \best{0.190$\pm$0.034} & \best{0.086$\pm$0.013} & 0.428$\pm$0.035\\
    \textbf{(10 scenes)} & \textbf{GS3LAM (Ours)}& 12.970$\pm$0.518  & 0.142$\pm$0.018 & 0.135$\pm$0.008 & \second{0.504$\pm$0.028}\\
    \textbf{} & \textbf{2DGS (Ours)} & \best{13.550$\pm$0.555}  & 0.174$\pm$0.016 & \second{0.087$\pm$0.012} & \best{0.509$\pm$0.025}\\
    \textbf{} & \textbf{SGS-SLAM (Ours)} &  \second{13.188$\pm$0.536}  & \best{0.190$\pm$0.022} & 0.094$\pm$0.005 & 0.500$\pm$0.025\\
    \bottomrule
\end{tabular}
}
}
\label{table:nvs-metrics-all}
\vspace{-3mm}
\end{table*}

\subsection{Novel-view Synthesis Performance}

Table~\ref{table:nvs-metrics-all} reports novel-view synthesis metrics across the three evaluation environments. Because PSNR, SSIM, and Depth L1 are computed only over fruit pixels rather than full frames, and because viewpoints are sparsely and actively selected rather than densely and uniformly sampled as in typical NVS benchmarks, the absolute PSNR (11.5--16.6~dB) and SSIM (0.13--0.33) values reported here are substantially lower than those commonly reported for 3D Gaussian Splatting \cite{kerbl3Dgaussians} on full , densely observed scenes. We therefore focus our analysis on the relative improvement introduced by our contributions rather than on these absolute values.

PSNR improves for every backbone in every environment when augmented with our robust mapping, and background pruning strategies, with gains ranging from +0.6~dB (2DGS, laboratory) to +4.9~dB (GS3LAM, simulation). MIoU shows the largest and most consistent improvement of all NVS metrics, increasing by 0.34--0.42 in simulation and up to 0.18 and 0.13 in the more challenging laboratory and real greenhouse environments, respectively, directly reflecting the benefit of the confidence-weighted semantic loss and background pruning for semantic map quality. SSIM improvements are more modest and less consistent: while all three backbones improve in simulation, gains in laboratory and real greenhouse settings are small and, for 2DGS, slightly negative.

Depth L1 shows a more mixed, backbone-dependent pattern. GS3LAM (Ours) exhibits a higher depth error than its vanilla counterpart in all three environments (+0.028 to +0.037~m), whereas 2DGS (Ours) improves slightly in every environment (-0.003 to -0.007~m). SGS-SLAM (Ours), our representative configuration, substantially reduces depth error in simulation (0.105~m to 0.052~m) but shows a small increase in the laboratory and real greenhouse settings (+0.008~m in both). We attribute this discrepancy primarily to the evaluation protocol rather than to an underlying loss of geometric accuracy: in simulation, Depth L1 is computed against noiseless, exact ground-truth depth, whereas in the laboratory and real greenhouse experiments the reference depth is itself obtained from a single noisy sensor capture at a novel viewpoint (Section~\ref{sec:experiments}), rather than from an independent, high-accuracy geometric measurement. Part of the reported depth error in these settings may therefore reflect sensor-to-sensor noise variance rather than true reconstruction inaccuracy, an evaluation limitation we revisit in Section~\ref{sec:discussion}.

Fig. \ref{fig:nvs-sgs-grid} presents novel views rendered by our proposed method and the corresponding baseline for each environment. Overall, compared with the vanilla GS models, our approach produces more accurate fruit masks, less noisy depth maps, and higher-quality plant reconstructions, while effectively suppressing background artifacts. However, in the real greenhouse environment, both the baselines and our method produce reconstructions that miss some fruits, particularly green peppers. This is likely because green peppers are more frequently misclassified by the fruit detector than ripe peppers, resulting in persistent semantic errors in the input observations. Although our framework is designed to mitigate the effects of segmentation errors, it cannot recover objects that are consistently assigned an incorrect semantic class by the perception model. This limitation highlights the distinction between robustness to noisy segmentation and robustness to systematic biases in the underlying semantic perception model.

\subsection{Robustness to Noise}

\begin{figure*}[htbp]
\centerline{\includegraphics[width=120mm]{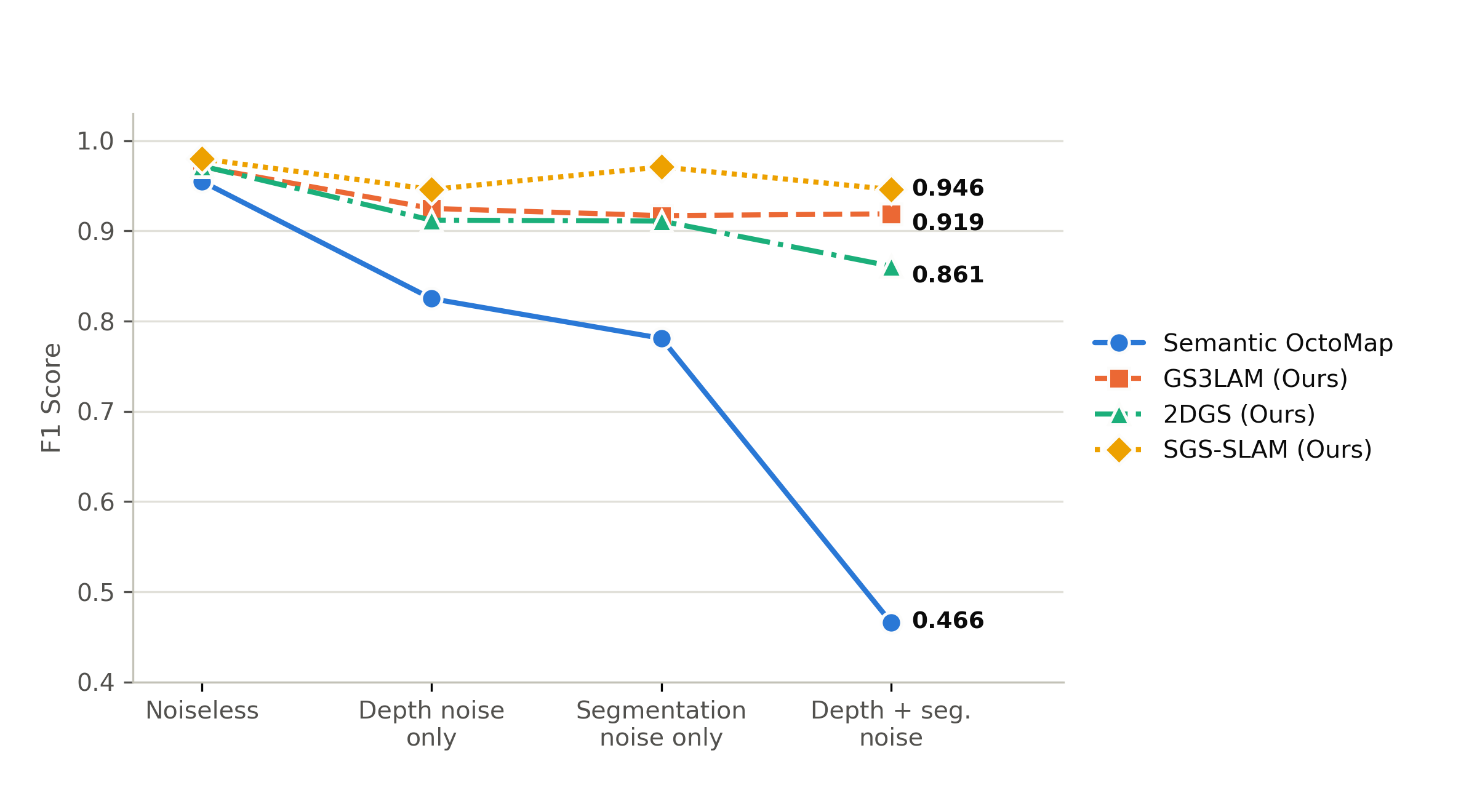}}
\vspace{-0.1cm}
\caption{F1 score of the Semantic OctoMap baseline and the three Gaussian Splatting backbones (augmented with our robust mapping and background pruning strategies) across the four noise conditions evaluated in the simulation environment. Semantic OctoMap degrades super-additively once both noise sources are present, while the Gaussian Splatting backbones remain comparatively stable. Full metrics are reported in Table~\ref{table:robustness-to-noise} (Appendix~\ref{sec:appendix-robustness}).}
\label{fig:robustness_noise_interaction}
\vspace{-0.1cm}
\end{figure*}

\begin{figure*}[h]
\centerline{\includegraphics[width=150mm]{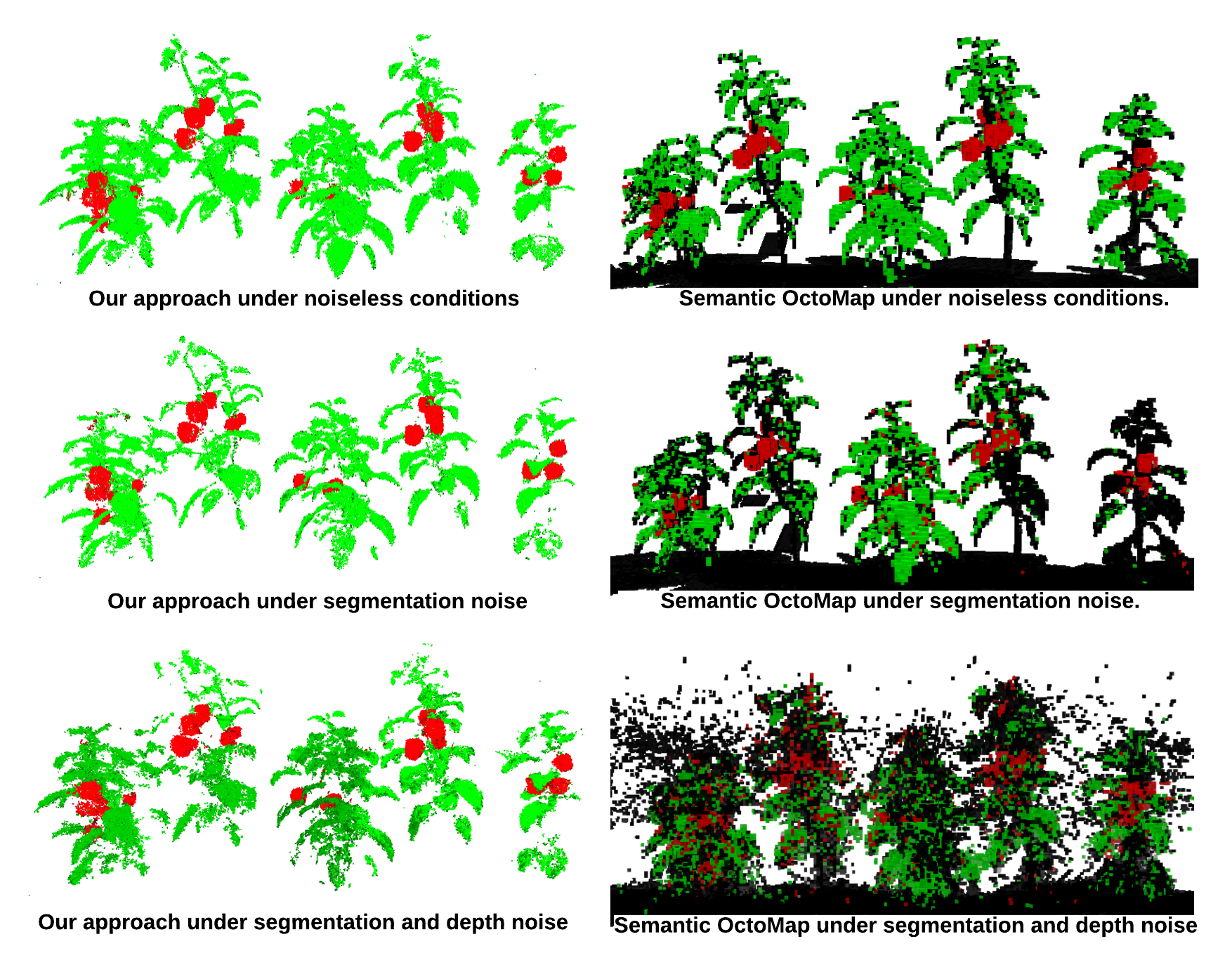}}
\caption{Sample reconstruction of a single crop row, demonstrating the robustness of the proposed approach (hybrid with SGS-SLAM backbone) to semantic segmentation and depth noise compared with a high-resolution OctoMap.}
\label{fig:octomap_vs_ours}
\vspace{-0.1cm}
\end{figure*} 

In this section, we evaluate the robustness of our framework to two sources of perception noise: semantic segmentation noise and depth noise. Using our simulation environment, we consider four experimental settings: (i) noiseless segmentation with noiseless depth, (ii) noiseless segmentation with noisy depth, (iii) noisy segmentation with noiseless depth, and (iv) noisy segmentation with noisy depth. Fig.~\ref{fig:robustness_noise_interaction} summarizes the F1 score of the Semantic OctoMap baseline and the three Gaussian Splatting backbones across these four conditions; the complete reconstruction and novel-view synthesis (NVS) metrics underlying this figure are reported in Table~\ref{table:robustness-to-noise} (Appendix~\ref{sec:appendix-robustness}).

As shown in Fig.~\ref{fig:robustness_noise_interaction}, Semantic OctoMap degrades sharply as noise is introduced, and the degradation is markedly super-additive: F1 drops from 0.955 (noiseless) to 0.825 under depth noise alone and 0.781 under segmentation noise alone, but collapses to 0.466 when both noise sources are present together, more than the sum of the two individual drops. Precision is hit hardest (0.963 to 0.346, Table~\ref{table:robustness-to-noise}), since each voxel's occupancy and class estimate is updated from whatever noisy evidence arrives, with no mechanism to reconcile conflicting depth and semantic cues across observations.

In contrast, all three Gaussian Splatting backbones augmented with our robust mapping, and background pruning strategies remain comparatively stable under the same combined noise condition, with F1 dropping only 3--11\% relative to their respective noiseless settings, compared with OctoMap's $-51\%$ (Fig.~\ref{fig:robustness_noise_interaction}). Moreover, the two noise sources affect Precision and Recall in a consistent and mechanistically interpretable way across all three backbones (Table~\ref{table:robustness-to-noise}): depth noise primarily reduces Precision (e.g., GS3LAM (Ours): 0.964 to 0.871) by introducing spurious or mispositioned Gaussians during initialization, while segmentation noise primarily reduces Recall (e.g., GS3LAM (Ours): 0.977 to 0.870), which in our case corresponds to fruit coverage.

Among the three backbones, SGS-SLAM (Ours) is the most resilient to segmentation noise specifically, with only a 0.026 drop in Recall compared with 0.107--0.112 for the other backbones, consistent with its overall best performance in Table~\ref{table:reconstruction-metrics-simulation}. Conversely, 2DGS (Ours) is the most sensitive to combined noise among the three backbones, in line with the Recall trade-off already observed for this backbone in Table~\ref{table:reconstruction-metrics-simulation}. Interestingly, NVS quality is comparatively insensitive to noise: PSNR for SGS-SLAM (Ours) stays within a narrow 16.0--16.6~dB range across all four conditions, indicating that noise primarily degrades geometric and semantic accuracy rather than photometric rendering quality. One exception is GS3LAM (Ours), whose Depth L1 nearly doubles under segmentation noise alone (0.069~m to 0.135~m) while remaining almost unchanged under depth noise alone (0.069~m to 0.068~m); we hypothesize that its jointly trained semantic classifier is more disrupted by label noise than the other backbones, indirectly degrading foreground/background convergence and, in turn, geometric accuracy, though we do not verify this mechanism directly. Full NVS metrics for all four conditions are reported in Table~\ref{table:robustness-to-noise} (Appendix~\ref{sec:appendix-robustness}).

Fig.~\ref{fig:octomap_vs_ours} illustrates the gap between Semantic Octomap and our 3DGS representation for one simulated crop row under combined noise. Semantic OctoMap produces visibly noisier maps than our approach, since it assumes a fixed noise model for all observations and cannot jointly reconcile depth and semantic evidence, causing individual voxels to converge to incorrect semantic values.

\subsection{Ablation Study}
To isolate the individual contribution of each component of our method, we conduct an ablation study using SGS-SLAM, our representative backbone, under the combined depth and segmentation noise condition used throughout this section. Starting from the vanilla SGS-SLAM baseline, we add robust mapping (RM) and background pruning (BP) both individually and jointly, and report the resulting reconstruction accuracy and NVS metrics in Table~\ref{table:ablation}.

\begin{table*}[htbp]
\centering
\caption{Ablation study of our Robust Mapping (RM) and Background Pruning (BP) strategies using SGS-SLAM under combined depth and segmentation noise in simulation. Results are averaged over 30 scenes.}
{
\resizebox{0.95\textwidth}{!}{
\begin{tabular}{ l | c c c c | c c c c}
    \toprule
     &  & Reconstruction Metrics &  &  &   & Novel View Synthesis Metrics &  & \\
    \midrule
    \textbf{Method}& \textbf{Chamfer}& \textbf{Precision$\uparrow$} & \textbf{Recall$\uparrow$} & \textbf{F1 Score$\uparrow$} & \textbf{PSNR[dB]$\uparrow$}& \textbf{SSIM$\uparrow$} & \textbf{Depth L1[m]$\downarrow$} & \textbf{MIOU$\uparrow$}\\
     & \textbf{Distance$\downarrow$ [m]}& &  &  &   &  &  & \\
    \midrule
    \textbf{SGS-SLAM (vanilla)} &0.076$\pm$0.028  & 0.783$\pm$0.092 & 0.778$\pm$0.053 & 0.777$\pm$0.063 & 12.258$\pm$0.847  & 0.189$\pm$0.024 & 0.105$\pm$0.017 & 0.294$\pm$0.045\\
     \textbf{SGS-SLAM+RM} & 0.029$\pm$0.012  & 0.882$\pm$0.026 & 0.938$\pm$0.047 & 0.909$\pm$0.032 & 15.068$\pm$0.950  & 0.227$\pm$0.027 & 0.064$\pm$0.014 & 0.590$\pm$0.038\\
    \textbf{SGS-SLAM+BP} & 0.034$\pm$0.019  & 0.943$\pm$0.068 & 0.501$\pm$0.106 & 0.647$\pm$0.096 & 15.830$\pm$0.948  & 0.227$\pm$0.036 & 0.188$\pm$0.037 & 0.519$\pm$0.038\\
    \textbf{SGS-SLAM+RM+BP(SGS-SLAM(Ours)) }& 0.013$\pm$0.002  & 0.936$\pm$0.019 & 0.958$\pm$0.026 & 0.946$\pm$0.016 & 16.156$\pm$0.993  & 0.238$\pm$0.032 & 0.052$\pm$0.012 & 0.713$\pm$0.049\\
    \bottomrule 
\end{tabular}
}
}
\label{table:ablation}
\vspace{-3mm}
\end{table*}

Table~\ref{table:ablation} shows that robust mapping and background pruning address distinct failure modes and are only effective when combined. Robust mapping alone (SGS-SLAM+RM) improves every metric relative to the vanilla baseline, most notably Recall (0.778 to 0.938) and MIoU (0.294 to 0.590, more than doubling), while improving Precision more modestly (0.783 to 0.882); this is consistent with robust mapping's role in correcting noisy Gaussian initialization and updates, which primarily governs how much of the true fruit and semantic content is recovered. Background pruning alone (SGS-SLAM+BP) shows the opposite trade-off: Precision reaches 0.943, the highest of any configuration including the full method, but Recall collapses to 0.501, below even the vanilla baseline (0.778), driving F1 down to 0.647, the worst of all four configurations. This indicates that, without robust mapping to first correct noisy or mispositioned Gaussians, the pruning criterion cannot reliably distinguish spurious background Gaussians from valid but noisy fruit Gaussians, and removes a large fraction of true fruit content along with background clutter. Background pruning alone also increases Depth L1 relative to the vanilla baseline (0.105 to 0.188~m); we attribute this to the removal of valid surface-covering Gaussians, which leaves gaps that are filled by less accurate remaining geometry during rendering.

Combining both components (SGS-SLAM+RM+BP, i.e., SGS-SLAM (Ours)) resolves this trade-off rather than merely averaging it: Precision reaches 0.936, nearly matching background pruning's standalone maximum (0.943), while Recall reaches 0.958, exceeding even robust mapping's standalone value (0.938). This indicates a synergistic rather than purely additive interaction: robust mapping first ensures that the Gaussians surviving pruning are accurately positioned and correctly classified, which in turn allows background pruning to remove spurious content aggressively without sacrificing valid fruit coverage. The combination also achieves the best Chamfer Distance (0.013~m, a further 55\% reduction over robust mapping alone) and the best Depth L1 (0.052~m), reversing the degradation observed when background pruning is applied in isolation. Overall, F1 improves from 0.777 (vanilla) to 0.946 (Ours), approaching the 0.980 achieved by SGS-SLAM (Ours) without any noise (Table~\ref{table:robustness-to-noise}, Appendix~\ref{sec:appendix-robustness}), and MIoU more than doubles (0.294 to 0.713), confirming that robust mapping and background pruning are both necessary and mutually reinforcing components of our framework.

\subsection{Phenotyping Performance}
In this section, we evaluate two task-oriented metrics that are particularly relevant to horticultural applications: fruit count accuracy and fruit volume accuracy, both of which are commonly used for phenotyping and yield estimation. We use the point clouds reconstructed by SGS-SLAM (Ours), our representative configuration (Table~\ref{table:reconstruction-metrics-simulation}), for this evaluation. These metrics are intended to demonstrate the practical applicability of our framework rather than to compare it against the baselines. Such a comparison would require point cloud denoising and clustering algorithms whose hyperparameters are highly sensitive to point cloud density, making it difficult to ensure a fair evaluation across different mapping methods.

We first extract fruit points from the reconstructed point cloud using the semantic labels and then cluster them using the DBSCAN algorithm. The \textit{fruit volume} of each cluster is estimated as the volume of the convex hull enclosing its 3D points. The fruit count and volume accuracies are computed as
\[
Accuracy=1-\frac{|gt_{\mathrm{value}}-estimated_{\mathrm{value}}|}{gt_{\mathrm{value}}}.
\]

The hyperparameters of the denoising and clustering algorithms are manually calibrated using the first crop row in the simulated environment and then evaluated on the remaining five rows to assess their generalization ability. Table~\ref{table:phenotyping-metrics} summarizes the resulting accuracy metrics; the reported Average is computed only over these five held-out rows (Rows 2--6), excluding Row 1, to reflect generalization rather than calibration performance.

\begin{table}[htbp]
\centering
\caption{Phenotypic metrics across six crop rows in simulation environment under noisy conditions, using SGS-SLAM (Ours) reconstructions. Row 1$^{*}$ was used to calibrate the denoising and clustering hyperparameters and is excluded from the reported average; Rows 2--6 are the held-out generalization set. Rows 5--6 correspond to the tomato plant variant.}
{
\resizebox{0.4\textwidth}{!}{
\begin{tabular}{c c c}
    \toprule
    \textbf{Crop Row} & \textbf{Fruit Count}& \textbf{Fruit Volume}\\
     & \textbf{Accuracy [\%]}& \textbf{Accuracy[\%]}\\
    \midrule
    \textbf{Row 1}$^{*}$ & 96.92$\pm$2.61& 79.98$\pm$15.42\\
    \textbf{Row 2} & 86.98$\pm$11.35& 88.00$\pm$9.33\\
    \textbf{Row 3} & 89.57$\pm$5.22& 87.19$\pm$5.41\\
    \textbf{Row 4} & 94.07$\pm$2.96& 73.37$\pm$6.67\\
    \textbf{Row 5} & 68.00$\pm$15.43& 75.45$\pm$16.90\\
    \textbf{Row 6} & 67.50$\pm$15.00& 73.52$\pm$13.26\\
    \midrule
    \textbf{Average}$^{\dagger}$ & 81.22& 79.51\\
    \bottomrule
\end{tabular}
}
}
{\raggedright\footnotesize \\$^{\dagger}$Computed over Rows 2--6 only.\par}
\label{table:phenotyping-metrics}
\vspace{-3mm}
\end{table}

The average fruit count and fruit volume accuracies are both close to 80\%, outperforming prior works that reported accuracies of approximately 60\% in similar simulated environments \cite{zaenker2021viewpoint, menon2023nbv}. This comparison is provided for reference only, as differences in the simulated environments, plant models, and experimental conditions may affect the reported accuracies and limit the validity of a direct comparison. Fruit count accuracy exhibits substantially greater variability across rows than fruit volume accuracy, ranging from 67.50\% to 94.07\% in the held-out rows, compared with 73.37\% to 88.00\% for fruit volume. This drop is concentrated in Rows 5 and 6 (68.00\% and 67.50\% count accuracy, respectively), which correspond to the tomato plant variant in our simulated scene. Unlike the bell pepper variants, tomato fruits often grow in tightly packed clusters where individual fruits are close to or touching one another, causing the DBSCAN step to more frequently merge multiple fruits into a single detected cluster. Such merges directly reduce the estimated fruit count, whereas the corresponding volume estimate is comparatively less affected, since the convex hull of a merged cluster still reasonably approximates the combined volume of the merged fruits.

\subsection{Runtime and Memory Usage}
\begin{figure}[htbp]
\centerline{\includegraphics[width=90mm]{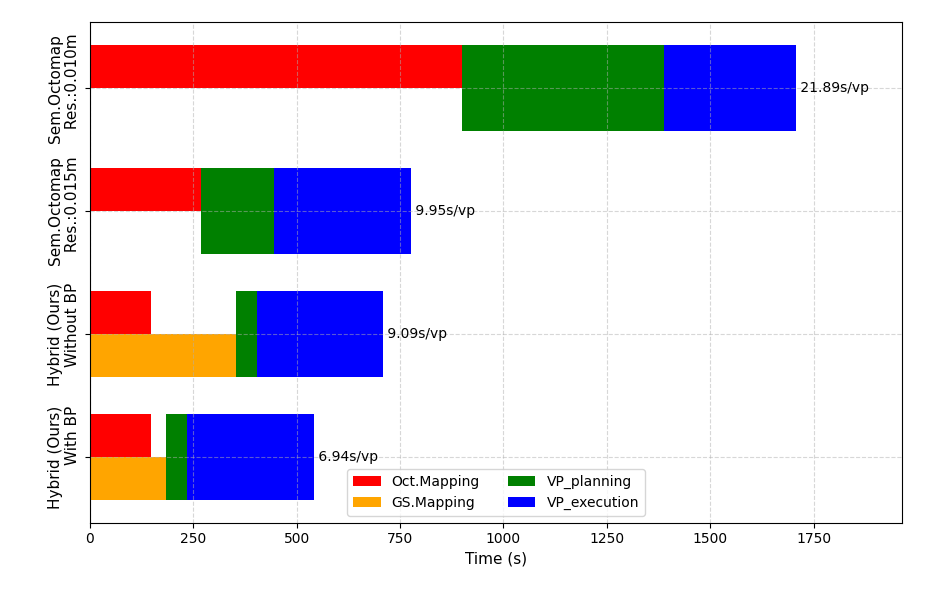}}
\caption{Runtime breakdown of our complete system, including OctoMap mapping, Gaussian Splatting mapping, viewpoint planning, and viewpoint execution. The OctoMap and Gaussian Splatting mapping modules run in parallel. Overall, our approach achieves an approximately threefold reduction in runtime compared with the high-resolution OctoMap baseline, primarily because the low-resolution OctoMap significantly reduces both mapping and viewpoint planning times.}
\label{fig:runtime}
\end{figure}

We decompose the total runtime into four main components: OctoMap mapping, 3D Gaussian Splatting mapping, viewpoint planning (including viewpoint sampling, filtering, evaluation, and graph-based planning), and viewpoint execution (including collision-free trajectory planning and trajectory execution). Figure~\ref{fig:runtime} reports the average runtime over 30 crop rows. In addition to the Semantic OctoMap baseline with a voxel resolution of 0.01~m, we also evaluate a coarser resolution of 0.015~m. Furthermore, we compare our hybrid framework with and without the proposed background pruning strategy.

Our approach is about three times as fast as the high-resolution (0.01~m) Semantic OctoMap baseline. This improvement stems from the fact that high-resolution OctoMaps require longer map update times and more expensive ray-casting during viewpoint evaluation than lower-resolution OctoMaps. Using a coarser voxel resolution of 0.015~m reduces the runtime of the baseline, although at the inherent expense of reconstruction quality.

Table~\ref{table:memory-usage} reports the number of Gaussians, peak GPU memory, and mapping runtime for each of the three Gaussian Splatting backbones. Unlike the full-system runtime decomposition in Fig.~\ref{fig:runtime}, the runtime column here isolates the 3DGS mapping component alone, measured per viewpoint under the offline evaluation protocol described in Section~\ref{sec:experiments}, and therefore excludes OctoMap mapping, viewpoint planning, and execution.

Background pruning reduces the number of Gaussians by 69--87\% across all three backbones (e.g., GS3LAM: 2.16M to 0.28M), directly confirming that the pruning criterion is aggressively and consistently removing background Gaussians regardless of the underlying backbone. This more compact representation also reduces mapping runtime by 42--52\%.

Peak GPU memory, however, does not scale proportionally with the number of Gaussians. 2DGS and SGS-SLAM show large memory reductions (57\% and 48\%, respectively) that roughly track their Gaussian count reductions, but GS3LAM's memory footprint drops by only 18\% despite having the largest reduction in Gaussian count (87\%) of the three backbones. This suggests that part of GS3LAM's memory footprint is independent of the number of Gaussians, plausibly stemming from its jointly trained semantic classifier network, the one architectural feature that distinguishes it from SGS-SLAM and 2DGS.

\begin{table}[htbp]
\centering
\caption{Memory usage and runtime performance in the Simulation Environment under noisy conditions}
{
\resizebox{0.5\textwidth}{!}{
\begin{tabular}{c c c c}
    \toprule
    \textbf{Method} & \textbf{Number of Gaussians$\downarrow$}& \textbf{GPU peak memory$\downarrow$} & \textbf{Runtime$\downarrow$}\\
    \textbf{} & \textbf{[M]}& \textbf{[GB]} & \textbf{[s/vp]}\\
    \midrule
    \textbf{GS3LAM} & 2.16 $\pm$ 0.11  & 1.75 $\pm$0.08 & 4.59$\pm$0.20\\
    \textbf{2DGS}& 2.29 $\pm$ 0.30  & 4.00 $\pm$0.53 & 4.23$\pm$0.41\\
    \textbf{SGS-SLAM}& 1.48 $\pm$ 0.13  & 3.20 $\pm$0.25 & 4.52$\pm$0.26\\
    \textbf{GS3LAM (Ours)}& 0.28 $\pm$ 0.09  & 1.43 $\pm$0.14 & 2.22$\pm$0.20\\
    \textbf{2DGS (Ours)}& 0.70 $\pm$ 0.22  & 1.70 $\pm$0.39 & 2.45$\pm$0.42\\
    \textbf{SGS-SLAM (Ours)}& 0.34 $\pm$ 0.06  & 1.67$\pm$0.08 & 2.37$\pm$0.17\\
    \bottomrule
\end{tabular}
}
}
\label{table:memory-usage}
\vspace{-3mm}
\end{table}

%% file: sections/discussion.tex
Our results show that among the evaluated Gaussian splatting backbones, SGS-SLAM achieved the best overall performance in terms of geometric reconstruction accuracy and robustness to segmentation noise. This result was somewhat unexpected, as 2DGS has demonstrated superior performance in other domains due to its direct surface optimization formulation. We believe that this outcome, as well as the high variance observed across backbones for certain metrics, is primarily attributable to differences in hyperparameter tuning rather than to inherent characteristics of the underlying methods. Each backbone involves several hyperparameters that require careful optimization, particularly those controlling Gaussian densification and pruning. Although we tuned these parameters to achieve a reasonable balance between densification and pruning for each backbone, we still observed substantial differences in the density of the resulting point clouds. Despite these inconsistencies, the proposed robust mapping and background pruning strategies consistently improves reconstruction accuracy, semantic quality, and computational efficiency, regardless of which underlying backbone is used. This backbone-agnostic behavior is an important result in its own right: the gains reported throughout Section~\ref{sec:results} are not an artifact of a single architecture, but a consequence of the noise-handling and pruning mechanisms themselves.

Across environments, the real greenhouse environment is consistently the most challenging across all metrics, reflecting the combined effects of wind-induced plant motion, uncontrolled illumination, sensor noise, and imperfect camera poses. In contrast, the controlled laboratory environment often matches or even outperforms the simulated environment on structural metrics such as Depth L1, SSIM, and MIoU, despite relying on real sensors. Simulation exhibits a clear advantage only in PSNR, likely because it does not account for wind-induced motion or camera pose errors, both of which introduce photometric inconsistencies in real-world observations.

Despite these strengths, our evaluation has an important limitation: quantitative geometric validation (Chamfer Distance, Precision, Recall, F1, and the robustness-to-noise and phenotyping analyses) is only possible in simulation, where exact geometric and semantic ground truth is available. Our real greenhouse experiments are limited both in sample size and, more fundamentally, in the availability of independent ground-truth geometry; the reference depth used for novel-view synthesis evaluation is itself a noisy sensor capture rather than an independent high-accuracy measurement (Section~\ref{sec:results}). This is precisely why we rely on simulation for controlled, quantitative validation of reconstruction accuracy and robustness, while treating our laboratory and real greenhouse experiments as evidence of practical applicability rather than as a substitute for ground-truth-based evaluation. Similarly, the segmentation noise model used in our study does not necessarily capture the systematic, spatially correlated failure modes of real segmentation models such as Detic and CrossGSD-Seg (e.g., confusion driven by occlusion, object size or atypical fruit coloration), and our robustness results should be interpreted with this simplification in mind.

Moreover, our framework assumes accurate camera pose estimation, obtained from the manipulator's forward kinematics and, in real-world experiments, a visual SLAM system. In earlier experiments, we explored enabling 3DGS-based camera tracking, similar to SGS-SLAM, but found that it degraded performance rather than improving it, because tracking relies on sufficiently dense and overlapping observations, which are not available under the sparse, actively selected viewpoints used in our framework. We therefore disabled tracking and instead relied on the accuracy of the underlying pose estimate. Camera pose error nonetheless remains an important challenge for real-world deployment, one that we did not explicitly quantify in this work. A promising direction for controlled environments is to incorporate fiducial markers to correct pose drift, as in our prior work~\cite{cuaran2025active}; extending this correction to unstructured, marker-free fields remains an open problem. 

Finally, our viewpoint sampling and exploration strategy leverages prior knowledge of row spacing and plant height to bootstrap mapping in trellised, row-structured horticultural environments. Applying our framework to other crop types or planting layouts (e.g., orchards or non-row field crops) would require adjusting or recalibrating these exploration priors. We note that active viewpoint planning is not the central contribution of this work; rather, it serves as a practical mechanism for generating the sparse, actively selected viewpoints under which we demonstrate that our hybrid semantic mapping approach remains accurate and efficient. Adapting the exploration strategy itself to a broader range of crop geometries is a direction for future work.

%% file: sections/conclusion.tex
We presented an approach for active semantic mapping in horticultural environments that combines a low-resolution octomap with a 3D Gaussian Splatting representation. Our method significantly improves reconstruction accuracy, enabling estimation of phenotyping traits such as fruit volume and fruit count with accuracies approaching 80\%. A carefully designed loss function along with a background pruning strategy enhances robustness to segmentation and depth noise; our ablation study shows that these two strategies address distinct failure modes and are most effective when combined, rather than simply adding their individual gains. These improvements hold consistently across three different Gaussian Splatting backbones, indicating that they stem from the noise-handling and pruning mechanisms themselves rather than from a particular architecture. Moreover, our approach is not resolution-dependent and runs considerably faster than high-resolution octomaps. As discussed in Section~\ref{sec:discussion}, several challenges remain, particularly camera pose accuracy in unstructured fields and the lack of ground truth geometry for validation beyond simulation. Nonetheless, we found that 3DGS-based mapping shows great promise for reconstructing horticultural environments and can be seamlessly integrated into robotic frameworks, providing a powerful tool for high-throughput phenotyping.

%% file: sections/appendix.tex
\subsection{Sample viewpoints with semantics}
\label{sec:appendix-sample-viewpoints-with-semantics}
\begin{figure*}[htbp]
\centerline{\includegraphics[width=180mm]{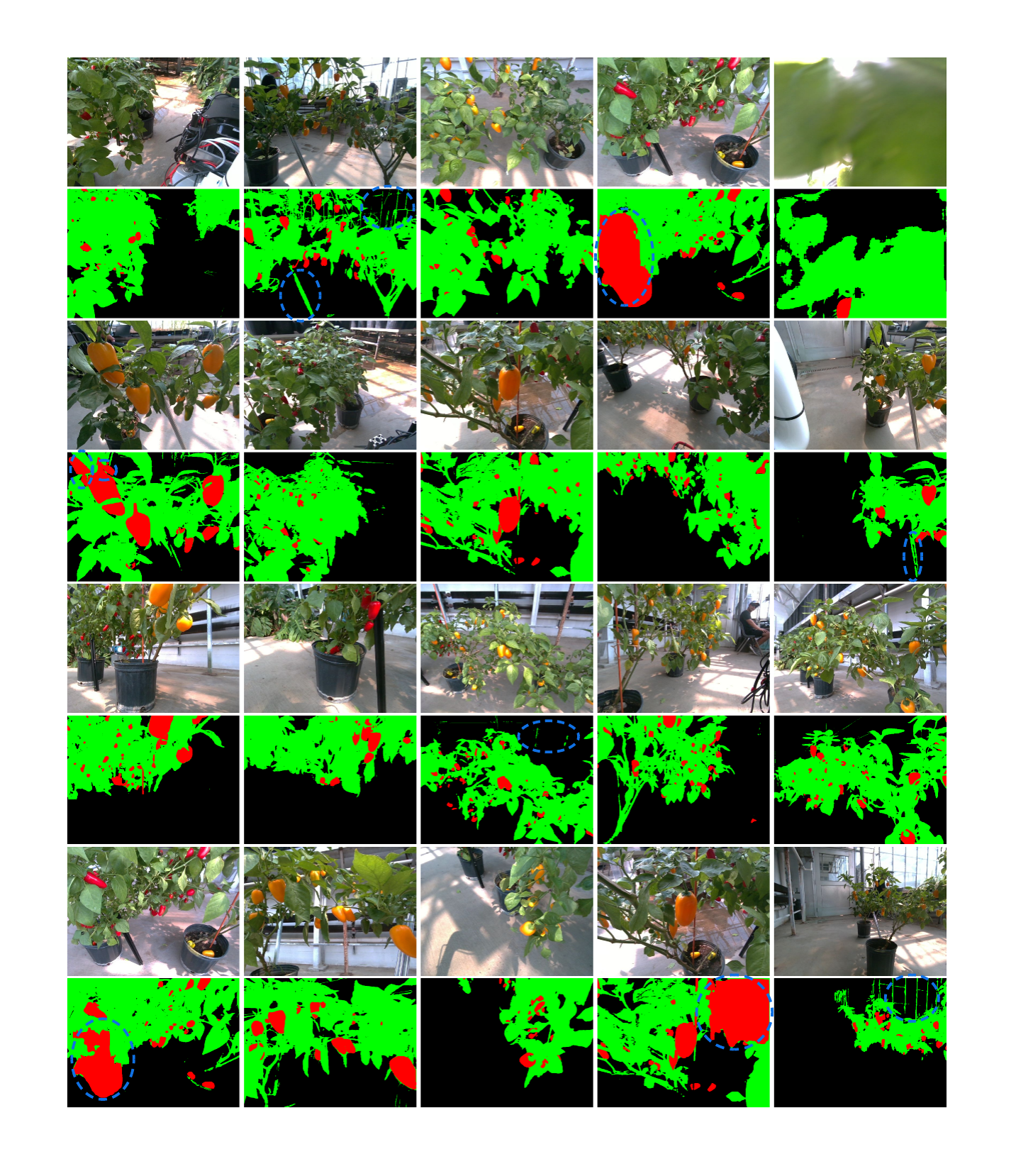}}
\vspace{-0.2cm}
\caption{Sample viewpoints captured using our active mapping approach in the real greenhouse environment. The corresponding semantic labels are shown below each RGB image. Semantic noise is highlighted by blue ellipses.}
\vspace{-0.1cm}
\label{fig:sample_rgb_and_semantics}
\vspace{-14pt}
\end{figure*}

Fig. \ref{fig:sample_rgb_and_semantics} shows representative images captured using our mapping approach. Note that the corresponding semantic predictions are far from perfect, often showing misclassified pixels.

\subsection{Sample Reconstructions}
\label{sec:appendix-sample-reconstructions}
\begin{figure*}[htbp]
\centerline{\includegraphics[width=180mm]{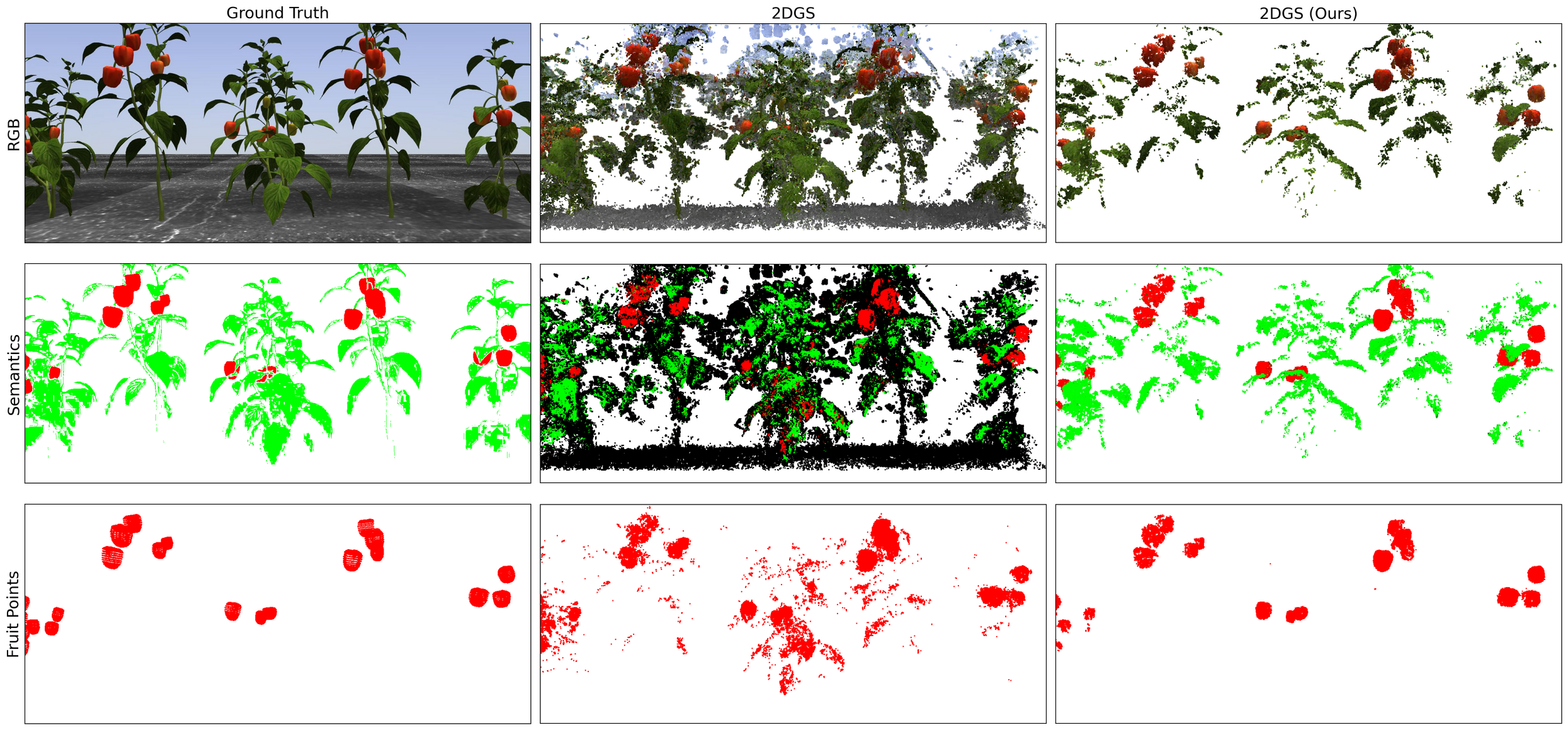}}
\vspace{-0.2cm}
\caption{Sample reconstructions of a bell pepper row from our simulation environment. Our approach produces cleaner and less noisy rgb, semantic and fruit point clouds than the 2DGS baseline.}
\vspace{-0.1cm}
\label{fig:2dgs_simulation_outputs_row1}
\vspace{-14pt}
\end{figure*}

\begin{figure*}[htbp]
\centerline{\includegraphics[width=180mm]{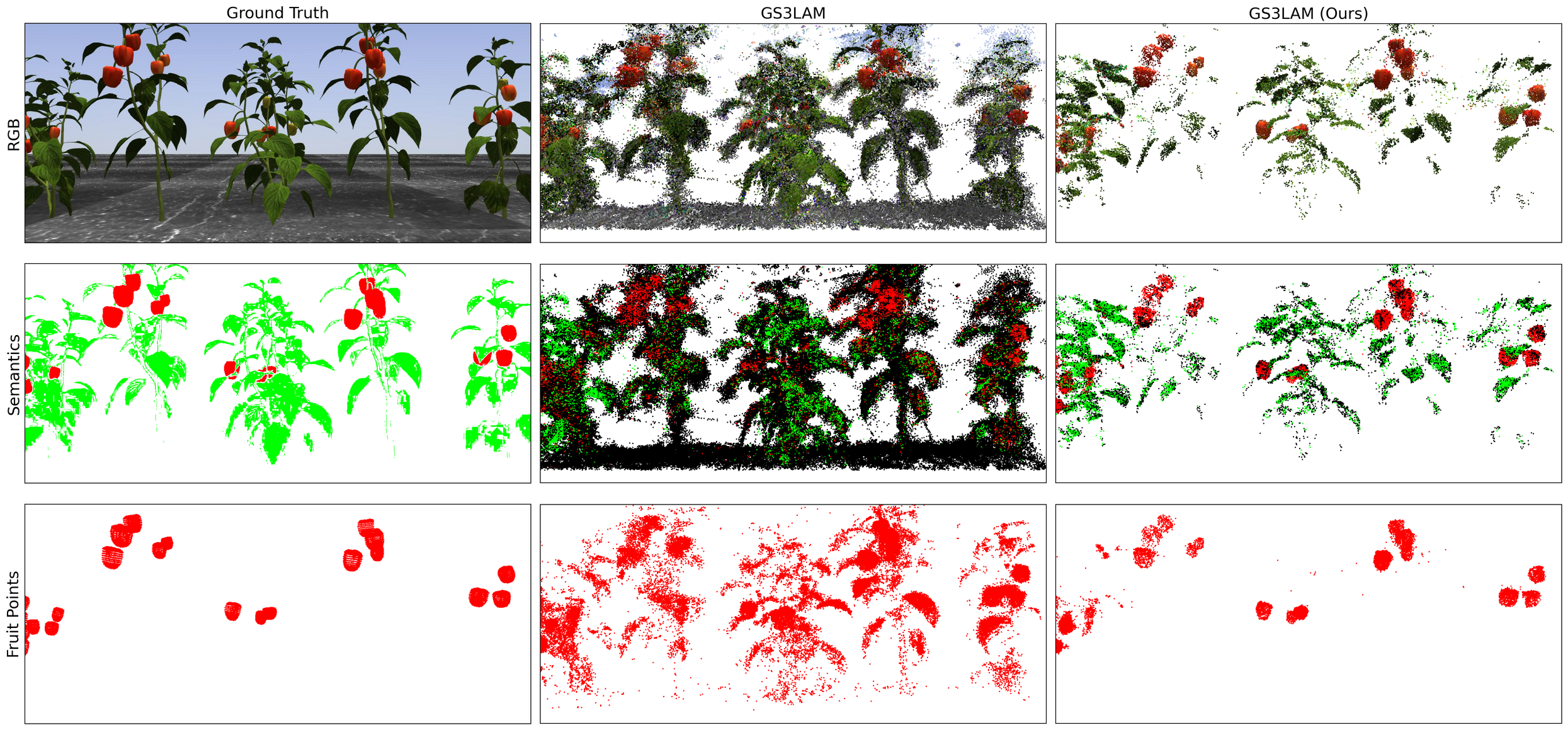}}
\vspace{-0.2cm}
\caption{Sample reconstructions of a bell pepper row from our simulation environment. Our approach produces cleaner and less noisy rgb, semantic and fruit point clouds than the GS3LAM baseline.}
\vspace{-0.1cm}
\label{fig:gs3lam_simulation_outputs_row1}
\vspace{-14pt}
\end{figure*}

Fig. \ref{fig:2dgs_simulation_outputs_row1} and Fig. \ref{fig:gs3lam_simulation_outputs_row1} show the color, semantic and fruit point clouds obtained with 2DGS and GS3LAM, for one of the bell-pepper rows in the simulation environment. 

\subsection{Novel View Synthesis Results}
\label{sec:appendix-nvs}
\begin{figure*}[htbp]
\centerline{\includegraphics[width=160mm, height=200mm]{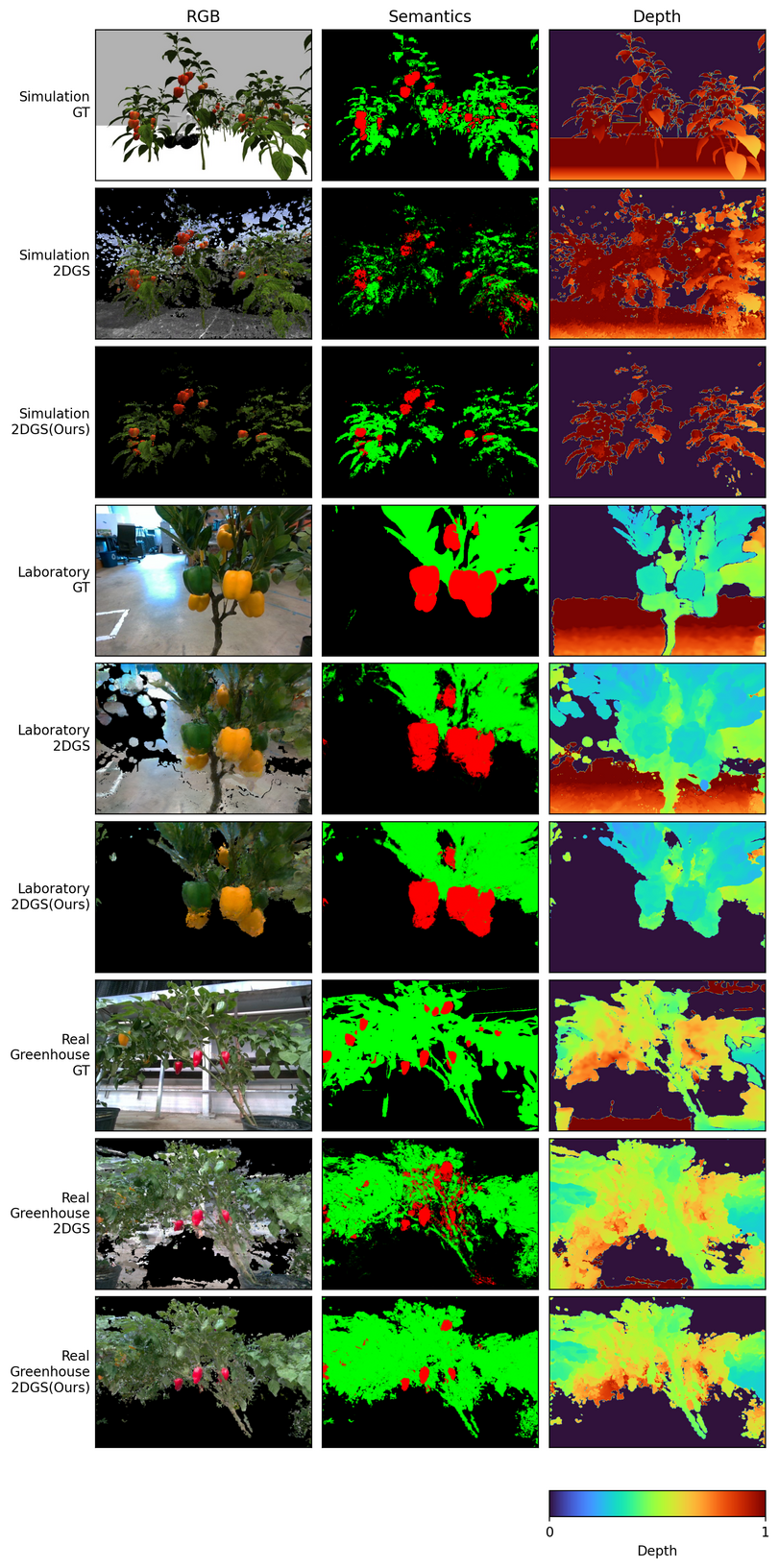}}
\vspace{-0.2cm}
\caption{Novel views generated using 2DGS and our approach across different environments.}
\vspace{-0.1cm}
\label{fig:nvs-2dgs-grid}
\vspace{-14pt}
\end{figure*}

\begin{figure*}[htbp]
\centerline{\includegraphics[width=160mm, height=200mm]{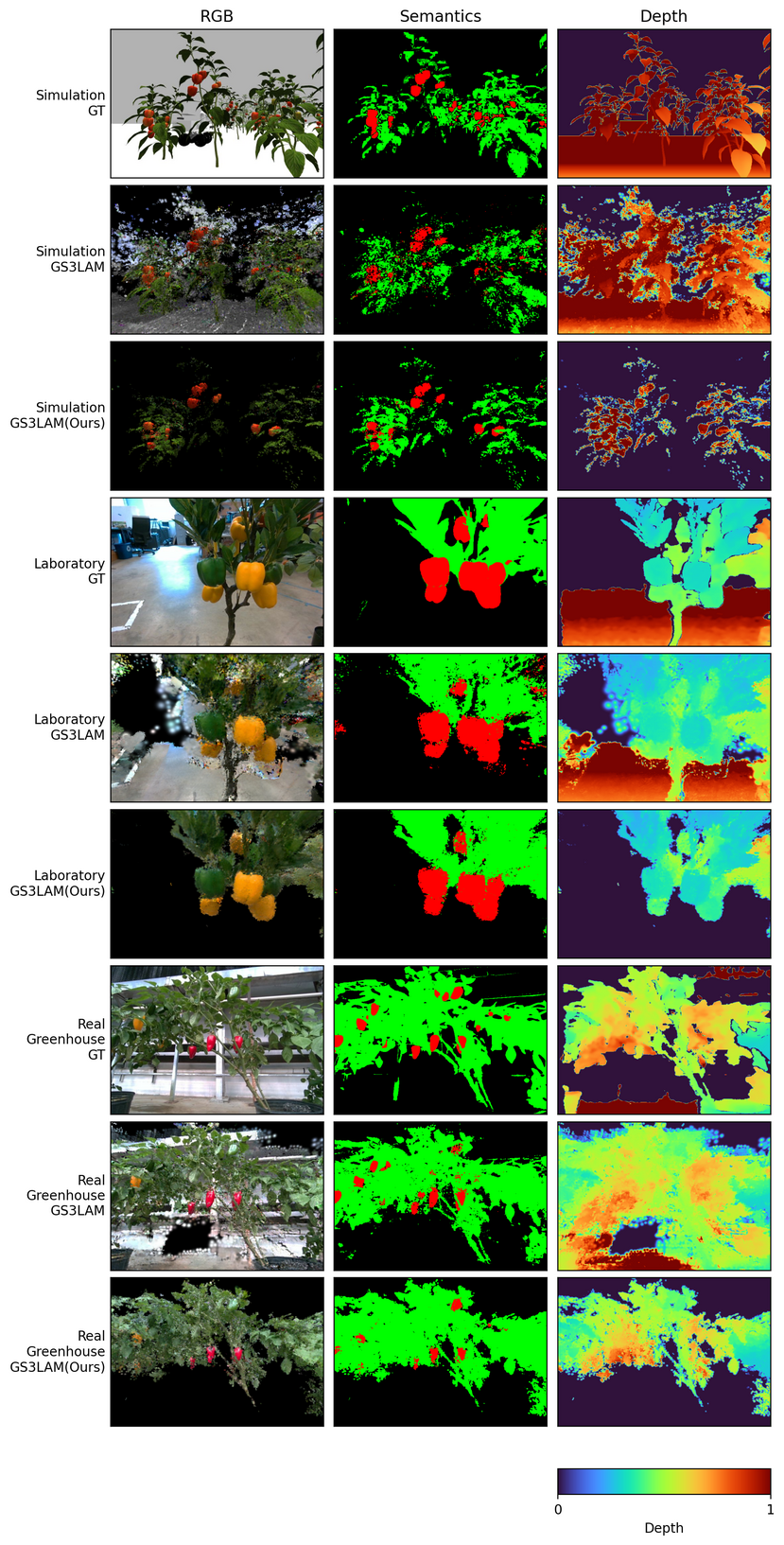}}
\vspace{-0.2cm}
\caption{Novel views generated using GS3LAM and our approach in different environments.}
\vspace{-0.1cm}
\label{fig:nvs-gs3lam-grid}
\vspace{-14pt}
\end{figure*}

Fig. \ref{fig:nvs-2dgs-grid} and Fig. \ref{fig:nvs-gs3lam-grid} show the color, semantic and depth images rendered with 2DGS and GS3LAM for unseen viewpoints in our simulated, laboratory and real-greenhouse environments.

\subsection{Robustness to Noise: Full Metrics}
\label{sec:appendix-robustness}
Table~\ref{table:robustness-to-noise} reports the full reconstruction and novel-view synthesis (NVS) metrics underlying the F1-score trends summarized in Fig.~\ref{fig:robustness_noise_interaction} (Section~\ref{sec:results}), for the Semantic OctoMap baseline and the three Gaussian Splatting backbones (augmented with our robust mapping and background pruning strategies) across all four noise conditions considered in this work.

\begin{table*}[htbp]
\centering
\caption{Robustness to depth and segmentation noise in the simulated environment. Results are averaged over 30 scenes.}
{
\resizebox{0.95\textwidth}{!}{
\begin{tabular}{c | c | c | c c c c | c c c c}
    \toprule
     & & &  & Reconstruction Metrics &  &  &   & Novel View Synthesis Metrics &  & \\
    \midrule
    \textbf{Method} & Seg. & Depth & \textbf{Chamfer}& \textbf{Precision$\uparrow$} & \textbf{Recall$\uparrow$} & \textbf{F1 Score$\uparrow$} & \textbf{PSNR[dB]$\uparrow$}& \textbf{SSIM$\uparrow$} & \textbf{Depth L1[m]$\downarrow$} & \textbf{MIOU$\uparrow$}\\
     & Noise & Noise & \textbf{Distance$\downarrow$ [m]}& &  &  &   &  &  & \\
    \midrule
    \textbf{Sem.Octomap} & x & x & 0.013$\pm$0.002  & 0.963$\pm$0.011 & 0.947$\pm$0.034 & 0.955$\pm$0.022 & NA  & NA & NA & NA\\
     & x & \checkmark & 0.023$\pm$0.002  & 0.766$\pm$0.021 & 0.888$\pm$0.094 & 0.825$\pm$0.039 & NA  & NA & NA & NA\\
    & \checkmark & x & 0.058$\pm$0.022  & 0.699$\pm$0.102 & 0.895$\pm$0.066 & 0.781$\pm$0.078 & NA  & NA & NA & NA\\
    & \checkmark & \checkmark& 0.101$\pm$0.025  & 0.346$\pm$0.127 & 0.740$\pm$0.214 & 0.466$\pm$0.155 & NA  & NA & NA & NA\\

    \midrule
    \textbf{GS3LAM (Ours)} & x & x & 0.017$\pm$0.018  & 0.964$\pm$0.018 & 0.977$\pm$0.015 & 0.971$\pm$0.015 & 17.366$\pm$1.225  & 0.298$\pm$0.043 & 0.069$\pm$0.021 & 0.792$\pm$0.033\\
     & x & \checkmark & 0.026$\pm$0.011  & 0.871$\pm$0.049 & 0.987$\pm$0.010 & 0.925$\pm$0.031 & 16.974$\pm$1.131  & 0.269$\pm$0.041 & 0.068$\pm$0.019 & 0.766$\pm$0.034\\
    & \checkmark & x & 0.021$\pm$0.016  & 0.972$\pm$0.019 & 0.870$\pm$0.053 & 0.917$\pm$0.033 & 16.727$\pm$1.298  & 0.239$\pm$0.047 & 0.135$\pm$0.027 & 0.698$\pm$0.031\\
    & \checkmark & \checkmark& 0.024$\pm$0.011  & 0.906$\pm$0.042 & 0.933$\pm$0.030 & 0.919$\pm$0.029 & 16.559$\pm$1.231  & 0.222$\pm$0.040 & 0.116$\pm$0.021 & 0.692$\pm$0.030\\

    \midrule
    \textbf{2DGS (Ours)} & x & x & 0.010$\pm$0.004  & 0.972$\pm$0.025 & 0.971$\pm$0.019 & 0.972$\pm$0.020 & 17.398$\pm$1.302  & 0.296$\pm$0.044 & 0.038$\pm$0.012 & 0.783$\pm$0.057\\
     & x & \checkmark & 0.021$\pm$0.011  & 0.880$\pm$0.062 & 0.948$\pm$0.047 & 0.912$\pm$0.054 & 16.678$\pm$0.929  & 0.250$\pm$0.037 & 0.054$\pm$0.018 & 0.709$\pm$0.065\\
    & \checkmark & x & 0.015$\pm$0.007  & 0.970$\pm$0.033 & 0.859$\pm$0.062 & 0.911$\pm$0.046 & 16.738$\pm$1.223  & 0.245$\pm$0.040 & 0.059$\pm$0.019 & 0.697$\pm$0.052\\
    & \checkmark & \checkmark& 0.023$\pm$0.011  & 0.907$\pm$0.054 & 0.822$\pm$0.089 & 0.861$\pm$0.071 & 16.348$\pm$1.018  & 0.219$\pm$0.038 & 0.070$\pm$0.024 & 0.642$\pm$0.057\\
    \midrule
    \textbf{SGS-SLAM (Ours)} & x & x & 0.023$\pm$0.040  & 0.968$\pm$0.018 & 0.993$\pm$0.006 & 0.980$\pm$0.011 & 16.561$\pm$1.008  & 0.281$\pm$0.036 & 0.035$\pm$0.01 & 0.778$\pm$0.048\\
     & x & \checkmark & 0.014$\pm$0.003  & 0.907$\pm$0.026 & 0.989$\pm$0.009 & 0.946$\pm$0.018 & 16.045$\pm$0.944  & 0.253$\pm$0.037 & 0.046$\pm$0.013 & 0.732$\pm$0.051\\
    & \checkmark & x & 0.022$\pm$0.032  & 0.976$\pm$0.016 & 0.967$\pm$0.028 & 0.971$\pm$0.017 & 16.611$\pm$1.076  & 0.263$\pm$0.034 & 0.039$\pm$0.013 & 0.763$\pm$0.052\\
    & \checkmark & \checkmark& 0.013$\pm$0.002  & 0.936$\pm$0.019 & 0.958$\pm$0.026 & 0.946$\pm$0.016 & 16.156$\pm$0.993  & 0.238$\pm$0.032 & 0.052$\pm$0.012 & 0.713$\pm$0.049\\
    \bottomrule
\end{tabular}
}
}
\label{table:robustness-to-noise}
\vspace{-3mm}
\end{table*}